\documentclass{article}

\usepackage{arxiv}

\usepackage[utf8]{inputenc} 
\usepackage[T1]{fontenc}    
\usepackage{hyperref}       
\usepackage{url}            
\usepackage{booktabs}       
\usepackage{amsfonts}       
\usepackage{nicefrac}       
\usepackage{microtype}      
\usepackage{graphicx}
\usepackage{amsmath}
\usepackage{calc}
\usepackage{float}
\usepackage{subfig}
\usepackage{amsfonts}
\usepackage[usenames,dvipsnames,svgnames,table]{xcolor}
\usepackage{amsthm}
\usepackage[utf8]{inputenc}
\usepackage{listings}
\usepackage{xcolor}
\usepackage{appendix}
\definecolor{codegreen}{rgb}{0,0.6,0}
\definecolor{codegray}{rgb}{0.5,0.5,0.5}
\definecolor{codepurple}{rgb}{0.58,0,0.82}
\definecolor{backcolour}{rgb}{0.99,0.99,0.99}
\definecolor{codeblack}{rgb}{0.05,0.05,0.02}
\lstdefinestyle{mystyle}{
  frame=tb,
  language=Python,
  aboveskip=3mm,
  belowskip=3mm,
  showstringspaces=false,
  columns=flexible,
  basicstyle={\small\ttfamily},
  numbers=none,
  numberstyle=\tiny\color{gray},
  keywordstyle=\color{blue},
  commentstyle=\color{codegreen},
  stringstyle=\color{codepurple},
  breaklines=true,
  breakatwhitespace=true,
  tabsize=3
}
\graphicspath{ {figures/} }
\newcommand\ddfrac[2]{\frac{\displaystyle #1}{\displaystyle #2}}

\title{Physics Informed Deep Learning for Transport in Porous Media. Buckley Leverett Problem.}

\author{
  Cedric G. Fraces\thanks{Stanford University (www.tenokonda.com)---\emph{Tenokonda}} \\
  Department of Energy Resources Engineering\\
  Stanford University\\
  Stanford, CA 94305 \\
  \texttt{cfraces@stanford.edu} \\
    \And
 Adrien Papaioannou \\
  Tenokonda\\
  London, UK\\
  \texttt{adrien@tenokonda.com} \\
   \And
 Hamdi Tchelepi \\
  Department of Energy Resources Engineering\\
  Stanford University\\
  Stanford, CA 94305 \\
  \texttt{tchelepi@stanford.edu} \\
}

\begin{document}
\maketitle

\begin{abstract}
\subsection*{Objective}
We present a new hybrid machine-learning/physics-based approach to reservoir modeling. The model is a neural network that is jointly trained to match any available experimental data and respect governing physical laws. The approach is used as a new way to model and history-match flow and transport problems in porous media. 

\subsection*{Methods}
The methodology relies on a series of deep adversarial neural network architecture with physics-based regularization. The network is used to simulate the dynamic behavior of physical quantities (i.e. pressure, saturation, composition) subject to a set of governing laws (e.g. mass conservation, Darcy’s law) and corresponding boundary and initial conditions. A residual equation is formed from the governing partial-differential equation and used as part of the training. Derivatives of the estimated physical quantities are computed using deep learning libraries that feature automatic differentiation algorithms. This strategy conditions the neural network to honor physical principles. The model thus adapts to best fit the data provided while striving to respect the governing physical laws. From an optimization stand-point, the governing equations are used as a regularization/constraint term. This allows the model to avoid overfitting, by reducing the variance and permits extrapolation beyond the range of the training data including uncertainty implicitely derived from the distribution output of the generative adversarial networks.

\subsection*{Results, Observations}
The methodology is used to simulate a 2-phase immiscible transport problem (Buckley-Leverett). From a very limited dataset, the model learns the parameters of the governing equation (the fractional flow curve) and is able to provide an accurate physical solution, both in terms of shock and rarefaction. We demonstrate how this method can be applied in the context of a forward simulation for continuous problems and show to use the trained networks to create higher-dimension representation at no extra cost. The use of these models for the inverse problem (history-match) is also presented, where the model simultaneously learns the physical laws and determines key uncertainty subsurface parameters. An additional benefit of the approach is that it is highly scalable and can leverage different computing architectures, including CPUs, GPUs and distributed clusters.

\subsection*{Novel/Additive Information}
The proposed methodology is a simple and elegant way to instill physical knowledge to machine-learning algorithms. This alleviates the two most significant shortcomings of machine-learning algorithms: the requirement for large datasets and the reliability of extrapolation. The principles presented in this paper can be generalized in innumerable ways in the future and should lead to a new class of algorithms to solve both forward and inverse physical problems.
\end{abstract}


\section{Introduction}
Earth systems predictions largely rely upon the modeling of complex fluid dynamics in the porous media characterizing the subsurface. In order to support better decision making, engineers and geo-scientists have developed forecasting tools and algorithms that are based on various delineations of mass and energy conservation along with the empirical Darcy law characterizing flow in porous media. Models must combine physical principles and observations in order to make accurate prediction. It is therefore important to quantify these two features:
\begin{enumerate}
    \item The complexity of the model in terms of scales and physics they emulate
    \item Their ability to assimilate data
\end{enumerate}
Numerical reservoir simulation lies on one end of that spectrum as they can capture complex physics and multiscale effects (heterogeneities, multi-phase, compositional, thermal, chemical reactions, geomechanics,...). The resolution of large numerical systems involve time intensive compuations and it is not rare to encounter models with single runtime of several hours to days. On the other hand, decline curve analysis (DCA, \cite{Arps1945}) is based on very simple physics that can easily be calibrated to observed data. Because of the large degree of uncertainty characterizing energy systems, most predictive models are parametric and must be updated against observed data. This gives rise to two classes of problems:
\begin{enumerate}
    \item The forward (or inference) problem where the goal is to make prediction on the time evolution of a quantity of interest (QOI) -typically the production of a field- based on a trusted model
    \item The inverse (or identification) problem where the goal is to estimate the parameters of the model -the subsurface properties for example- that honor the observation on the QOI and a prior understanding we have of the system.
\end{enumerate}
 
Energy systems present a unique set of challenges to machine learning methods. They have large financial and environmental footprints, wide uncertainties and obey strict physical constraints. Recent advances in instrumentation, telemetry and data storage have allowed operators to rely more and more on data to make decisions. However, the integration of all this data represents a challenge, while the need to inform decisions in a timely manner has become an essential competitive differentiator in most industries, \cite{DeBezenac2017}. One of the main motivation for the present work is that to this day, we lack a robust methodology for the application of machine learning to Earth systems, \cite{Reichstein2019}. The challenges identified are:
\begin{itemize}
    \item Data assimilation
    \item Scale and Resolution
    \item Generalization outside training scope
    \item Interpretability causal discovery
    \item Physical consistency of governing physical laws
    \item Uncertainty Quantification
    \item Limited access to labeled data
\end{itemize}
In this work, we propose to address some of these issues. We will be leveraging some of the most recent advances in statistical and deep learning and more particularly the use of deep neural networks. We plan to utilize some of their known advantages: their ability to model complex data structures all while limiting some of their shortcomings: their inability to generalize well and to provide interpretable solutions. After a literature review of existing state of the art  physics-informed machine learning techniques, we will present our method and its application to the Buckley-Leverett problem \cite{BuckleyLeverett1942}. We will extend with variations of the original problem including multiple physics and dimensions. We finish with a discussion on variational and generative in the context of uncertainty quantification.

\section{Related work}
\label{sec:related_work}
\subsection{Data driven approaches}
In spite of a large interest and commercial needs, there is relatively little applied research on data driven solutions for reservoir engineering. Classical engineering tools with a proven track record (DCA, type curves, material balance, Capacitance and Resistance models, Analytical methods, numerical simulation) are being used by a vast majority of engineers and geoscientists. Attempts to leverage data and statistical approaches are relatively new and coincide with the rise of unconventional production in the US. These are characterized by a large increase in the number of wells drilled in a short time period combined with a limited understanding of the physics governing the behavior of shale formations. Due to the availability of vast amount of data, researchers and engineers have started experimenting with data-driven approaches to get insights into the production patterns. The present work was initiated with a method to generalize the concept of decline to cases where it normally fails (unconventional, secondary recovery, infill drilling). Mohaghegh et al.\ \cite{Mohaghegh2011} developed a top-down approach for modeling and history matching of shale production based on statistical and pattern recognition models and applied their approach to three shale reservoirs. Sun et al.\ \cite{Sun2018} applied Long-Short Term Memory algorithm (LSTM) to predict a well's oil, water, and gas production. Their work led to an improvement of the production forecast in comparison with standard DCA models. However, the tested portion of their time series exhibits weak variations in comparison with the training part and it remains unclear whether the algorithm can generalize well to complex time series. 

\subsection{Physics Informed}
Others have applied Physics Informed Neural Networks to the problem of single well production forecast. These results encouraged the expansion to the more complex problem of 2-Phase transport in 1D and then in more dimensions. Chiramonte (\cite{Chiarmaonte2018}) shows 2 examples of resolution of the Laplace equation in 2D. They do some error analysis. They compare the solution with traditional finite volume and show that the solution obtained using neural networks loses some resolution.
Rudy (\cite{Rudy2018}) presents a study on the data driven identification of parametric partial differential equations. They use multivariate sparse regression techniques (Lasso, Ridge) to calculate the coefficients of Burger and Navier-Stokes equations. Their approach allow the coefficients to have arbitrary time series, or spatial dependencies. Considering the large collection of candidate terms for constructing the partial differential equation, it is assumed that a sufficiently large library of 'base' functions is used to approximate the PDE's coefficient. The approach, although very promising suffers two major drawbacks: first, the use of numerical differentiation in their gradient descent optimization; second, the assumption of sparse representation and multivariate regression is valid only if the library of basis function is sufficiently rich. This reveals untrue or impractical for most cases. Zabaras, et al.\ (\cite{MoZabaras2018}, \cite{YinhaoZabaras2018} and \cite{YinhaoZabaras2019}) use a convolutional encoder decoder approach to build probabilistic surrogates for multiple gaussian realizations of a permeability field. Their models are enriched by a physics based loss computed using Sobol filters to mimic the differential forms of a PDE. This approach is similar to the one presented by Bin Dong et al.\ (\cite{Long2018_pdenet}) .Their approach is fast and treats the problem as one of image to image regressions. If the method shows promises for elliptic equations (flow), they lack the ability to extrapolate well especially in the hyperbolic transport problem. We draw inspiration from the work of Raissi and Perdikaris, \cite{RaissiJML2018}, \cite{raissi2017physicsI}, \cite{raissi2017physicsII} who present a new method that addresses the two issues aforementioned by assigning prior distributions in the form of artificial neural networks or Gaussian processes. Derivatives of the prior can now be evaluated at machine precision using symbolic or automatic differentiation (Deep learning libraries such as Tensorflow, \cite{tensorflow2015-whitepaper} are well suited for this type of calculation). This allows a certain noise level in the observations and removes the need to manually compute the derivatives of the solution in order to evaluate the residual of the PDE. Altough results are demonstrated for the Burger's equation in 1D with a sinusoidal initial condition, we show that it fails in the case of Buckley Leverett with a constant initial and boundary condition.

\subsection{A word on Automatic differentiation}
Automatic differentiation (AD) lies in between symbolic and numerical differentiation. If the method does provide values of the derivatives and keeps track of the derivative values of expressions (as opposed to their expression), it does so by establishing first symbolic rules for the differentiation. AD relies on the postulate that all derivation operations can be decomposed in a finite set of basic operations for which derivatives are known (example: $\frac{dx^2}{dx}=2x$). These basic operations can be combined through the so-called chain rule in order to recover the derivative of the original expression. This hybrid approach gets derivatives at machine precision in a time that is comparable to getting them manually (with a small overhead). \cite{Baydin2018_AD} provides a revue of AD methods and their application to machine learning. Open source software libraries for AD such as Tensorflow (\cite{tensorflow2015-whitepaper}) allow the creation and compilation of computational graphs that are then used to differentiate quantities of interest at observation points. Once computed, the graph is used through the entire optimization process (whether for training a neural network or for computing the residual of a PDE). This static implementation results in significant savings in terms of computational time.

\section{Methodology}
The method we present uses neural networks to solve partial differential equations (PDE) of the form:
\begin{equation}
    \mathcal{R}(u_t, t, x, u, u_x, u_{xx},\dots) = 0
\end{equation}
Similarly to what we do in classical numerical analysis, we assume a representation for the unknown solution $u$. Finite elements formulations assume that $u$ is represented by a linear combination of basis functions. 
\begin{equation}
    u \approx \sum\limits_{i} \hat{u}_i\phi_i
\end{equation}
In this approach instead, we assume that it is represented by function compositions with series of linear and non linear transformations. This can conveniently be represented by a feed forward multi-layer perceptron.
\begin{equation}
    u \approx g(\mathbf{W}_0\times(g(\mathbf{W}_1...)+\mathbf{b_0})
\end{equation}
Where $g$ is a nonlinear activation function (sigmoid, tanh, ReLu,...), $\mathbf{W}_i$ are weight matrices and $\mathbf{b_i}$ are bias vectors. As an example, a network with 2 hidden layers  and a tanh activation function and $\nu = (x,t)$ as an input is of the form:
\begin{equation}
    \hat u(\mathbf{\nu}, \mathbf W, \mathbf b) = \sum_i^{N_0}  \sum_j^{N_1}  \sum_k^{N_2}   tanh\left( W_{lk}^{[2]} \left[ tanh \left( W_{ki}^{[1]} \left[ tanh \left( W_{ij}^{[0]} \nu_j + b_i^{[0]} \right) \right] + b_k^{[1]} \right) \right] + b_l^{[2]} \right)
    \label{eq:MLP_2layers}
\end{equation}
$\hat{u}$ is a continuous representation. For any $x,t$ it will produce an output. It is also a differentiable representation. This means that we can compute any derivatives with respect to space and time. We can therefore construct the new quantity of interest $\mathcal{R}$ (for residual). $\mathcal{R}$ is a neural network that has the same architecture as $\hat{u}$, the same weights and bias, but different activation functions due to the differential operator. If the idea seems simple, several technical challenges need to be addressed in order to implement it. One of them is the difficulty to differentiate an equation of type \ref{eq:MLP_2layers} (more complex in reality) in a quick and efficient manner. We leverage the capabilities of modern software like Tensorflow (\cite{tensorflow2015-whitepaper})to compute the differential of this complex form automatically (using the same technique used in backpropagation). The software library's API allows to efficiently code the whole formulation in very few lines of code as shown in the following code snapshot.
\begin{lstlisting}[language=Python, caption=Network implementation of PINN for Buckley Leverett Problem]
import numpy as np
import tensorflow as tf

# Network for QOI (Saturation) as a function of (x,t)
def net_saturation(...):
    ...
    mlp = tf.layers.dense(x, mlp_config.layer_size_lst[i], 
                          activation=mlp_config.activation_lst[i],
                          kernel_initializer=tf.contrib.layers.xavier_initializer(),
                          name=mlp_config.main_name + '_layer_' + str(i), reuse=reuse)
    ...

# Network for PDE residual
def net_pde_residual(s, x, t):
    s_t = tf.gradients(s, t)[0]
    s_x = tf.gradients(s, x)[0]
    s_xx = tf.gradients(s_x, x)[0]
    lambda_swc = 0.0
    lambda_m = 2
    lambda_sor = 0.0
    nu = 0.001
    frac = tf.divide(tf.square(s - Swc), tf.square(s - lambda_swc) + tf.divide(tf.square(1 - s - lambda_sor), lambda_m))
    frac_s = tf.gradients(frac,s)[0]
    f = s_t + frac_s * s_x
    f = tf.identity(f, name='f_pred')
    return f

# Loss function
def discriminator_loss(logits_real, logits_fake):
    # x = logits, z = labels
    # tf.nn.sigmoid_cross_entropy_with_logits <=> z * -log(sigmoid(x)) + (1 - z) * -log(1 - sigmoid(x))
    dis_loss_real = tf.reduce_mean(tf.nn.sigmoid_cross_entropy_with_logits(logits=logits_real, labels=tf.zeros_like(logits_real)))
    dis_loss_fake = tf.reduce_mean(tf.nn.sigmoid_cross_entropy_with_logits(logits=logits_fake, labels=tf.ones_like(logits_fake)))
    dis_loss = dis_loss_real + dis_loss_fake
    return dis_loss

def generator_loss(logits_fake, logits_posterior, pde_residuals, w_posterior_loss, w_pde_loss):
    # x = logits, z = labels
    # tf.nn.sigmoid_cross_entropy_with_logits <=> z * -log(sigmoid(x)) + (1 - z) * -log(1 - sigmoid(x))
    gen_loss_entropy = tf.reduce_mean(logits_fake)
    gen_loss_posterior = tf.reduce_mean(tf.multiply((w_posterior_loss - 1.0), tf.nn.sigmoid_cross_entropy_with_logits(logits=logits_posterior, labels=tf.ones_like(logits_posterior))))
    gen_loss_pde = w_pde_loss * tf.reduce_mean(tf.square(pde_residuals), name='loss_pde_form')
    gen_loss = gen_loss_entropy + gen_loss_posterior + gen_loss_pde
    return gen_loss, gen_loss_entropy, gen_loss_posterior, gen_loss_pde
     
\end{lstlisting}

The weights of the two neural networks for \texttt{net\_saturation} and \texttt{net\_pde\_residual} are learned by minimizing a two elements loss function \texttt{discriminator\_loss} - \texttt{generator\_loss} defined as the cross entropy loss of the difference between observation and solution at the boundaries and initial conditions and evaluation of the residual at collocation points. Depending on the the amount of data available and information on the governing physics, two classes of problem emerge, namely the forward (or inference) and inverse (or identification) problem. These are addressed subsequently. Traditional numerical approaches based on space/time discretization require a high resolution in order to capture discontinuities near the shock region. On the other hand, higher order discretization schemes are more accurate but require more data and are more difficult to converge. They also do not extrapolate well outside of the training space. Raissi et al.\ proposed the idea to use deep neural networks as presented to solve the 1-D Burger equation \cite{raissi2017physicsI}, \cite{raissi2017physicsII} and the norm of the PDE as a regularizing term in the training process.

\subsection{On the properties of Neural Networks}
Training a neural network can be very tedious and finding general rules that apply through all networks can be tricky. Training remains an exercise that is very much an art more than a science. A set of heuristic rules have been proposed and are followed within the machine learning community to tune the hyper-parameters of a model. In general, the network's architecture (depth or number of layers and width or number of neurons and form of activation functions) relies on the modeler's choice of expressivity vs trainability. It is commonly admitted that deeper and wider networks are more expressive (meaning that they can capture more complex forms) but are harder to train (take more time to evaluate and optimize).

\subsubsection{Universal Approximation}
One of the most interesting properties of neural networks is their ability to interpolate any functional form with remarkable accuracy and stability. Hornik (\cite{HORNIK1991251}) demonstrates that a feed-forward network with a single hidden layer containing a finite number of neurons and arbitrary bounded and non-constant activation function can approximate any continuous function in compact subsets of $\mathbb{R}^n$. This property of neural networks is expressed in the universal approximation theorem and was first proposed in 1989. More recently, these results were expended to limit the width of the networks and showing that with $O(n)$ layers, networks could approximate a wide variety of functions when given appropriate parameters (\cite{NIPS2017_7203}).

\subsubsection{Regularization, smoothing}
Lipschitz regularization of deep neural networks is a property that ensures the robustness of these function to changes in the input. One metric to assess the robustness of neural networks to small perturbations is the Lipschitz constant, which establishes a relationship between input perturbation and output sensitivity. \cite{Henri_LipschitzReg} shows that by restricting the weights $\theta$ of the neural network $f_{\theta}$, it becomes a Lipschitz function i.e:
\begin{equation}
    \|f_{\theta}(x) - f_{\theta}(y)\| \leq L \|x - y\|
\end{equation}
Where $L$ is called the Lipschitz constant of the neural network. The determination of that constant even for a small network is a NP-hard problem but several techniques can estimate lower bounds. 
That regularization provides a smoothing effect to noisy input data and makes neural networks good candidates for data assimilation. It is also a convenient property when we utilize these networks as generators of solutions rather than classifier of regressors. 

\subsubsection{High dimensional expansion}
Recent studies have focused on explaining how networks learn \cite{IntriguingNN2014} shows that the representations learned depend on a large number of network weights rather than a few and that representations learned are very non linear. This second property makes it possible to fool networks with well-crafted but slight modifications of the inputs. This finding led to the work on adversarial networks that we will cover in a later section. Further work on neural network training and explainability \cite{DBLP_Shwartz-ZivT17} bridges the gap between deep learning and information theory \cite{Shannon_1948}. A neural network is a Markov chain of representations. Information can only decrease when crossing a Markov chain. Under any invertible mapping, mutual information does not change. This de-correlates information from computational complexity.  \cite{DBLP_Shwartz-ZivT17} shows that training occurs in two phases. The first phase is called empirical risk minimization (learn to fit the labelled data). The second phase is called the forgetting phase and corresponds to the elimination of the unimportant information for the label. This property is very valuable when we solve physical problems in 2D and 3D.

\section{Problem statement and Preliminary Results}
\subsection{Buckley-Leverett Problem}
\begin{figure}[H]
\begin{center}
\includegraphics[width=.5\linewidth]{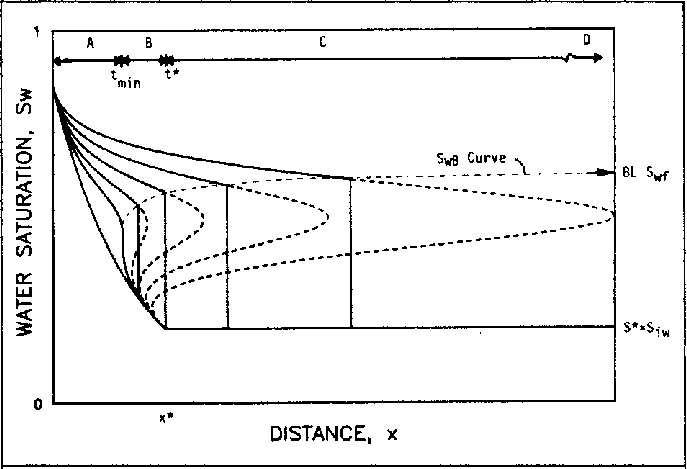}
\end{center}
   \caption{Example of Buckley-Leverett solutions. The full (non-physical) solution is marked by the dotted lines. The solid lines represent the physical solutions resolved by the shock.~\cite{mohsen1985modification}}
\label{fig:1}
\end{figure}

The Buckley-Leverett theory of immiscible displacement led to one of the most ubiquitous modeling techniques in the upstream oil and gas industry. It was first introduced in 1943 \cite{buckley1942mechanism} and it estimates the rate at which an injected water bank moves through a porous medium. The approach uses fractional flow theory and is based on the following assumptions: 
\begin{itemize}
    \item Incompressible fluids
    \item One dimensional, horizontal flow 
    \item Steady flow $Q_t = Q_w + Q_o = cst$
    \item Immiscible fluids (like oil and water)
    \item No capillary pressure $(P_w=P_o)$
\end{itemize}
The 1-D Buckley-Leverett equation is defined as ~\cite{buckley1942mechanism}:
\begin{equation}
    \label{eq:1}
    \frac{\partial S_w}{\partial t} + \frac{\partial}{\partial x} \left(\frac{Q}{\phi A}f_w(S_w)\right) = \phi \frac{\partial S_w}{\partial x}
\end{equation}
Where $S_w(x,t)$ is the water saturation, $Q$ is the total flow rate, $\phi$ is the porosity, $A$ is the cross sectional area, and $f_w(S_w)$ is the fractional flow function defined as:
\begin{equation}
    f_w = \frac{\lambda_w}{\lambda_T}\\
        = \frac{1}{1 + \frac{k_{ro}\mu_w}{k_{rw}\mu_o}}
\end{equation}
For the purposes of this problem we impose inlet and outlet water saturation ($S_{w, in}, S_{w, out}$). Depending on the fractional flow function, $f_w$,  the Buckley-Leverett equation can yield solutions that have very different forms. The solution is self similar and its general shape is conserved as it propagates through space and time. A strict adherence to the Buckley-Leverett equation can have results where a lower saturation has a slower speed than higher saturation counterparts~\cite{mohsen1985modification}. This non-physicality is resolved by a shock as shown in figure~\ref{fig:1}.\\

\subsection{Identification Problem}
\subsubsection{Formulation}
This formulation is also called "inverse problem" as we would like to determine the hidden parameters of the PDE that produce a solution compatible with the observed data. The solution $\hat{S}$ and the nonlinear equation defining the residual $\mathcal{R}$ are approximated using two deep dense neural networks. The residual of the PDE defines the physics constraint to the problem.

\begin{equation}
\label{eq:Buckley_residual}
    \mathcal{R} = \frac{\partial \hat{S}}{\partial t} + \frac{\partial f}{\partial S}\frac{\partial \hat{S}}{\partial x}
\end{equation}.
Where the fractional flow $f$ is a nonlinear equation defined as:
\begin{equation}
    \label{eq:frac_flow}
    f(S) = \frac{(S - \tilde{S}_{wc})^2}{(S - \tilde{S}_{wc})^2 + (1 - S - \tilde{S}_{or})^2/\tilde{M}}
\end{equation}

In Eq.~\ref{eq:frac_flow} the $\tilde{.}$ terms are learn-able parameters which enable the learning of the governing differential equation. Perdikaris et al.\ present a solution on the Burger's equation by training a deep neural network with a loss term ($\mathcal{L}$) defined by eq.~\ref{eq:Loss_overall}.

\begin{equation}
    \label{eq:Loss_overall}
    \mathcal{L} = \mathcal{L}_s + \omega \mathcal{L}_r
\end{equation}

Where $\mathcal{L}_s$ is the observed data error defined by eq.~\ref{eq:function_loss} and $\omega$ is a hyper-parameter introduced to control the importance of the residual loss over the data loss.

\begin{equation}
    \label{eq:function_loss}
    \mathcal{L}_s = \frac{1}{N}\sum_{i=1}^N |\hat{S}(t^i,x^i) - S^i|^2
\end{equation}

And $\mathcal{L}_r$ is the error with respect to the governing differential equation defined by eq.~\ref{eq:residual_loss}.

\begin{equation}
    \label{eq:residual_loss}
    \mathcal{L}_r = \frac{1}{N}\sum_{i=1}^N |\mathcal{R}(t^i,x^i)|^2
\end{equation}

\begin{figure}[H]
\begin{center}
\includegraphics[width=0.8\linewidth]{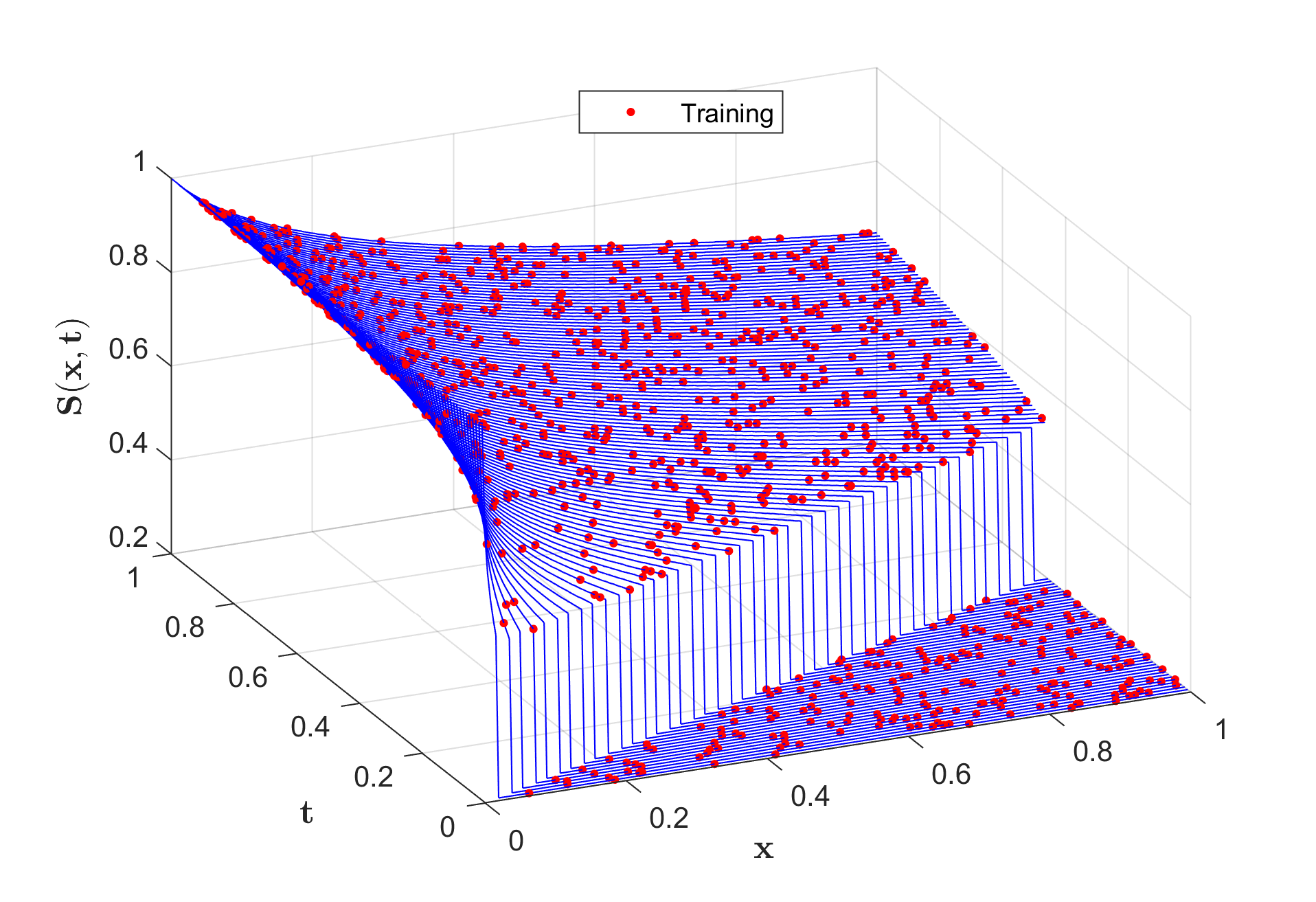}
\end{center}
   \caption{Training data sampling for the Buckley Leverett solution. Blue lines are the actual solutions at different time steps while red dots are training data measurements (1000 points).}
\label{fig:discovery_random_sampling}
\end{figure}

\begin{figure}[H]
\begin{center}
\includegraphics[width=1\linewidth]{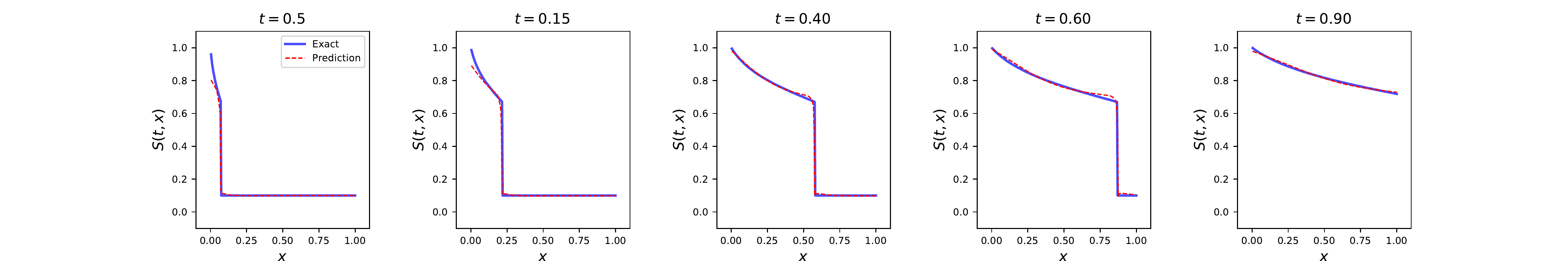}
\end{center}
   \caption{Result of matching the solution of Buckley-Leverett equation given $x,t,S$ within the domain and with moving parameters (discovery problem). The true solution (blue) and the one computed using a neural network (red) are overlaid at five different time steps $(0.05, 0.15, 0.4, 0.6, 0.9)$}
\label{fig:BL_solution_1D}
\end{figure}

We apply the method described by Raissi et al.\ to the Buckley-Leverett equation (1-D). We use 1000 training data points ($N=1000$ in eq.~\ref{eq:function_loss} and \ref{eq:residual_loss}). Training data is shown in figure~\ref{fig:discovery_random_sampling} while a series of results are shown in figure~\ref{fig:BL_solution_1D}. 


\subsubsection{Various sampling schemes and error analysis}
\begin{figure}%
    \centering
    \subfloat{\includegraphics[width=0.8\linewidth]{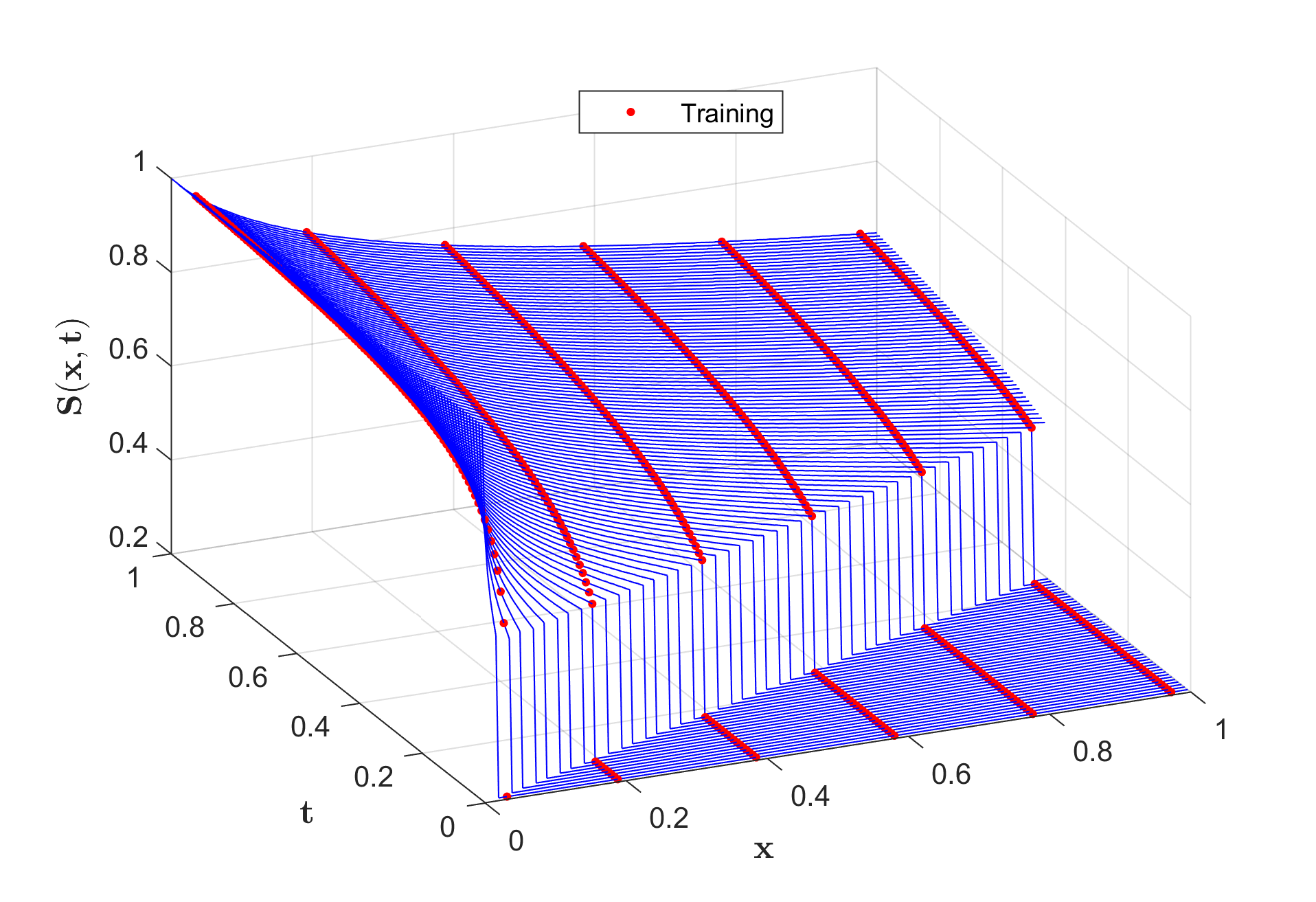}}
    \qquad
    \subfloat{\includegraphics[width=1\linewidth]{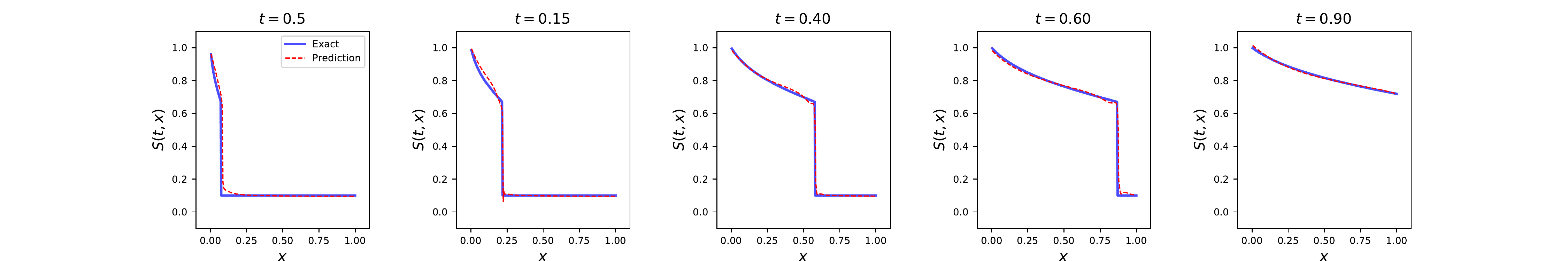}}
    \caption{Training sample (top) and predictions results in the problem of identification with fixed sampling locations}%
    \label{fig:discovery_fixed}%
\end{figure}

\begin{figure}%
    \centering
    \subfloat{\includegraphics[width=0.8\linewidth]{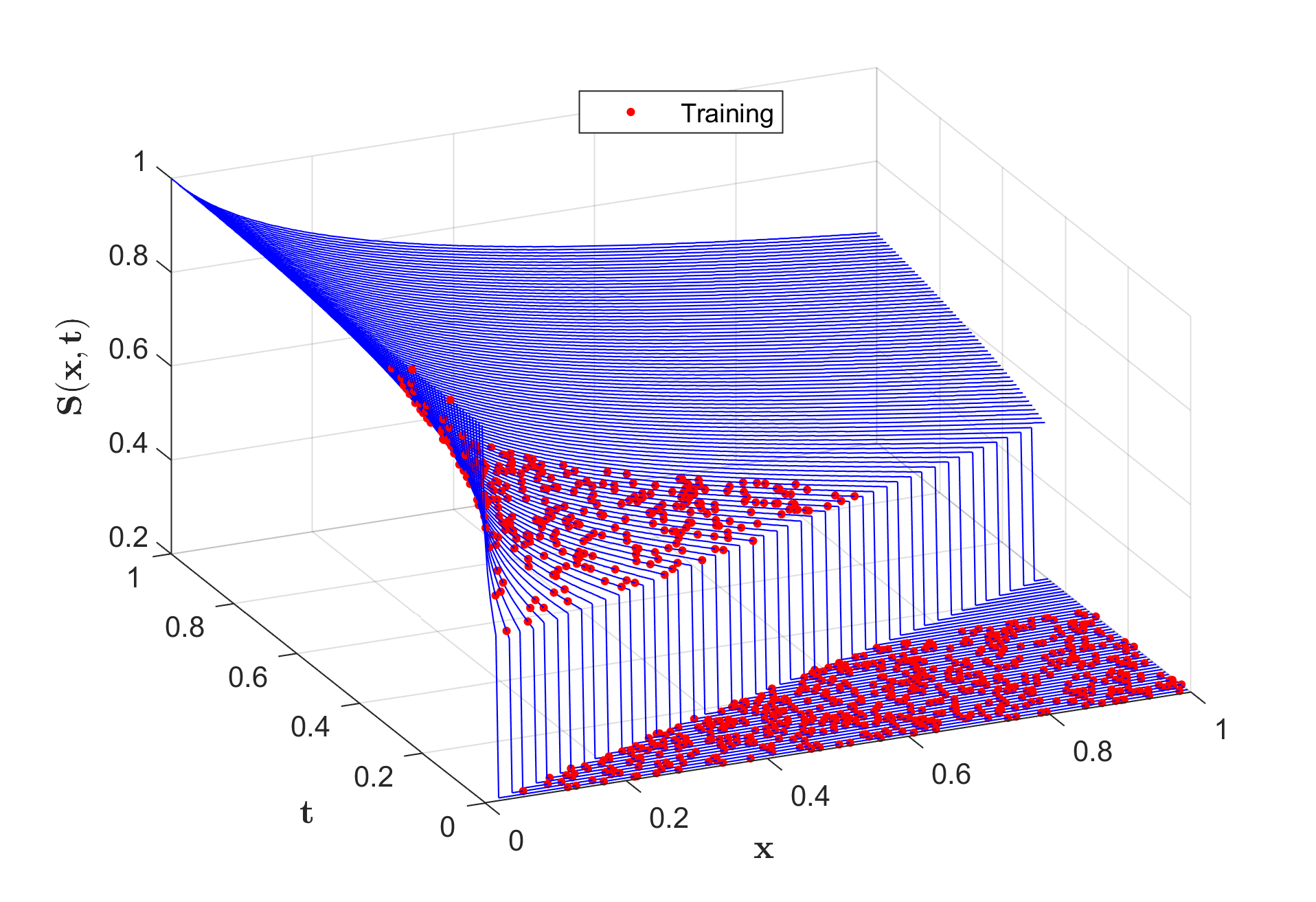}}
    \qquad
    \subfloat{\includegraphics[width=1\linewidth]{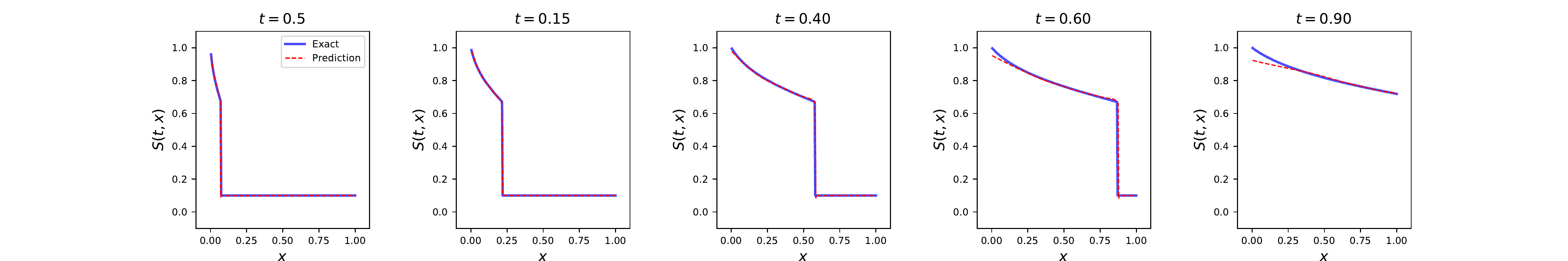}}
    \caption{Training sample (top) and predictions results in the problem of identification with sampling points at early times only}%
    \label{fig:discovery_early}%
\end{figure}

We evaluate the sensitivity of the inverse nethod to the nature of the data sampled. Instead of a random sampling, we force the data to be measured at specific locations (emulating the presence of multiple observation wells around an injector. The results are presented in figure~\ref{fig:discovery_fixed}.
We then implement a sampling strategy that considers data only at early times (one third of total simulation time). Results are presented in figure~\ref{fig:discovery_early}.

\begin{figure}[H]
\begin{center}
\includegraphics[width=0.7\linewidth]{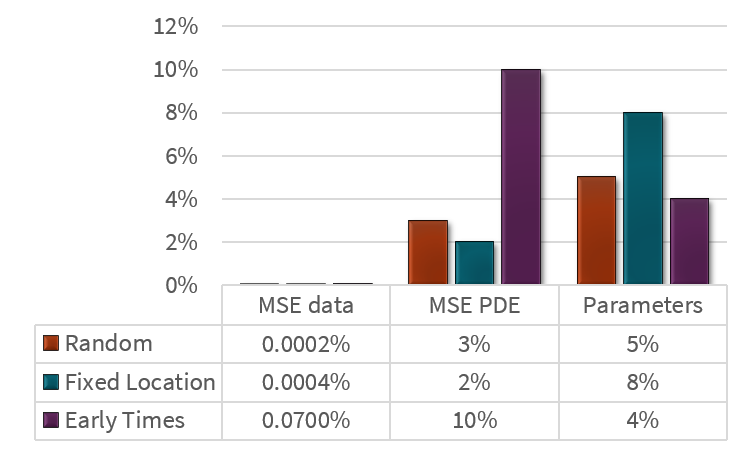}
\end{center}
   \caption{Error analysis for various solution sampling strategies in the discovery problem. Each colored box represents the entire Buckley Leverett solution. The relative mean square error for the overall saturation (MSE data), the residual PDE (MSE PDE) and the $\lambda$'s parameters.}
\label{fig:BL_discovery_error}
\end{figure}

The error on the various parameters depends on the number and location of the training data. We distinguish three types of error in this problem:
\begin{enumerate}
    \item The error on the data : $\frac{1}{N_{tot}}\sum_{i=1}^{N_{tot}} |\hat{S}(t^i,x^i) - S^i|^2$
    \item The error on the residual of the PDE : $\frac{1}{N_{tot}}\sum_{i=1}^{N_{tot}} |\mathcal{R}(t^i,x^i)|^2$
    \item The error on the parameters of the PDE : $\frac{1}{N_{tot}}\sum_{i=1}^{N_{tot}} \|\hat{\lambda_i}-\lambda_i\|$
\end{enumerate}
Where $N_{tot}$ represents the resolution of the original solution.
The data error asymptotically converges to 0 when the number of training points increases and covers the whole time/space domain. Figure~\ref{fig:BL_discovery_error} shows how the discovery of parameters problem performs well under different training data-sets.
These results confirm the intuition that neural networks offer a good alternative for data assimilation. These results demonstrate that assessing the parameters of a Buckley-Leverett equation from data can be achieved provided that enough training data is available.
We notice that the error on the PDE tends to increase when a large portion of the dataset is not used for training. This problem becomes critical in cases where only boundary data and initial are used for training (Inference problem). Section \ref{sec:inference} addresses this issue along with the solutions proposed to alleviate it.

\subsection{Inference problem}
\label{sec:inference}
The inference problem consists in finding the forward solution of a PDE considering a limited set of observations located at the boundaries and initial conditions. Note that this is the minimum required number of auxiliary conditions to establish a unique solution to a first order PDE with two variables. 
\begin{figure}%
    \centering
    \subfloat{\includegraphics[width=0.8\linewidth]{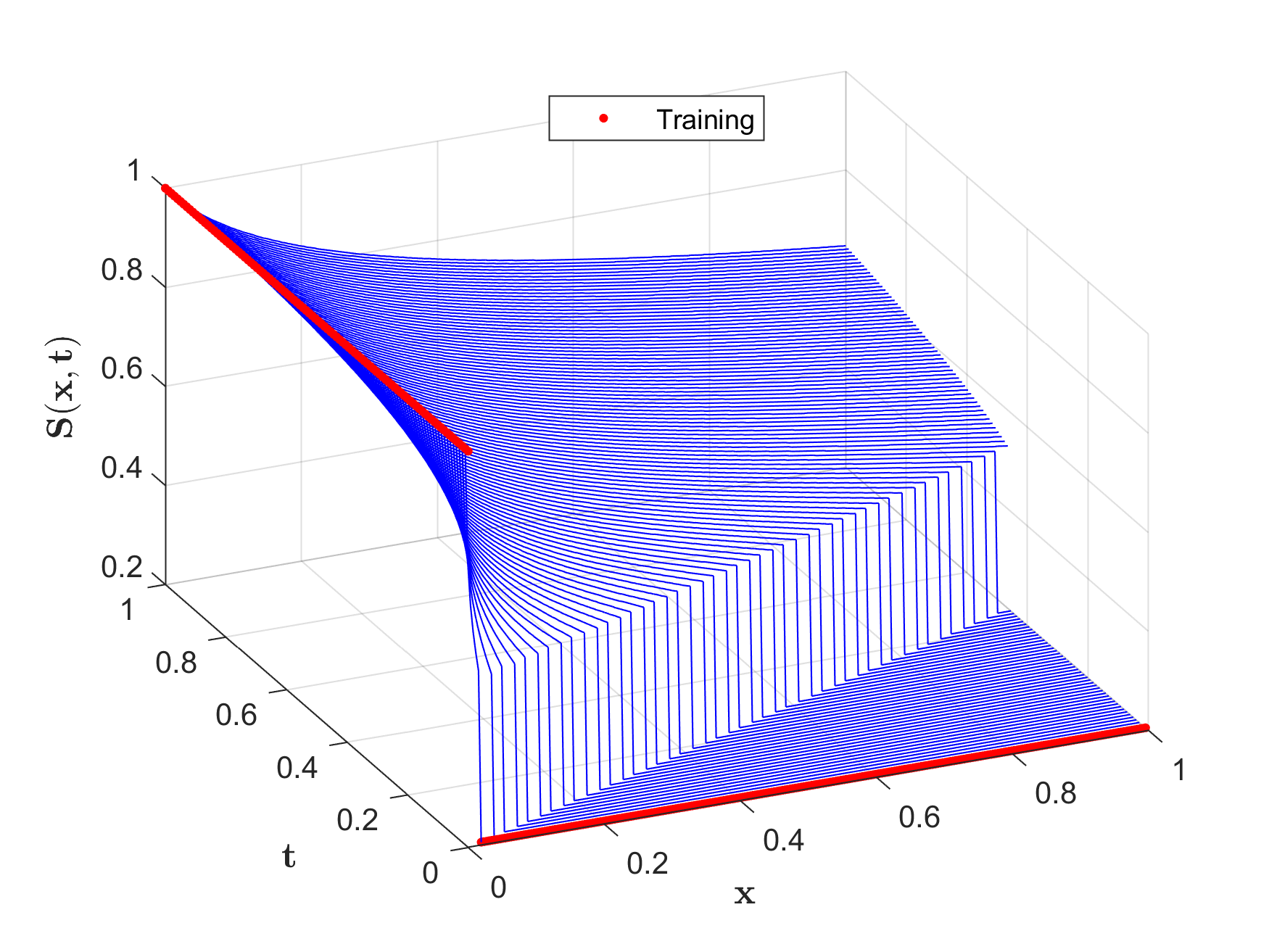}}
    \qquad
    \subfloat{\includegraphics[width=1\linewidth]{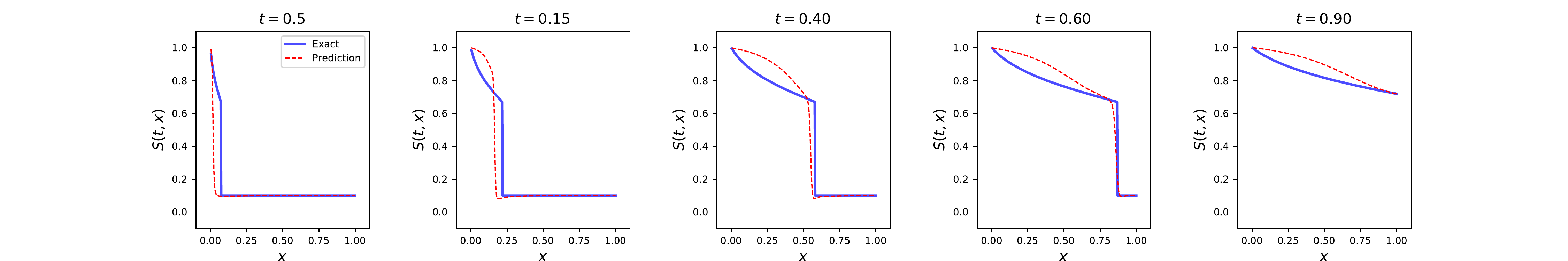}}
    \caption{Training sample (top) and predictions results in the problem of inference with sampling points at initial and boundary conditions only}%
    \label{fig:inference_fail}%
\end{figure}
The overall formulation of the problem is very similar to what was presented in the discovery section except that the loss function $\mathcal{L}$ now features three least square components:
\begin{equation}
    \mathcal{L} = \mathcal{L}_{0} + \mathcal{L}_{b} + \mathcal{L}_r
\end{equation}
Where the terms of the loss are defined by:
\begin{gather}
    \mathcal{L}_0 = \frac{1}{N_0}\sum\limits_{i=1}^{N_0} \|\hat{S}(0,x^i) - S_0^i\|^2\\
    \mathcal{L}_b = \frac{1}{N_b}\sum\limits_{i=1}^{N_b} \|S(t^i,x_{min}) - \hat{S}(t^i,x_{min})\|^2 + \|S(t^i,x_{max}) - \hat{S}(t^i,x_{max})\|^2\\
    \mathcal{L}_r = \frac{1}{N_f}\sum\limits_{i=1}^{N_r} \|\mathcal{R}(t^i,x^i)\|^2
\end{gather}
$\mathcal{L}_0$ denotes the loss on the initial conditions, $\mathcal{L}_b$ the loss on the boundary conditions and $\mathcal{L}_r$ the loss on the residual which penalizes the solution for not respecting the governing equation at all collocation points. Typically, we rely on a much larger number of collocation points ($N_r$) than points at the boundary ($N_b$) and initial conditions ($N_0$).
\begin{equation}
    N_r >> N_b + N_0
\end{equation}
When we add derivatives of higher orders to the residual expression, the form of the residual and the order of the derivatives determine the number of boundary conditions needed to have a well posed problem. Hence there is a relationship between the order of the problem and the type of data/loss functions we feed to the training problem.
The inference problem is more complex to resolve than the identification one. A vanilla implementation with two networks training simultaneously to minimize the loss function leads to poor results as shown in figure~\ref{fig:inference_fail}. We explore two options that help the training. The first one is \textbf{transfer learning}, the second one is \textbf{generative models}. We briefly explain the concept of transfer learning and will expand on generative models in a later section.

Transfer learning is the improvement of learning in a new task using knowledge learned from a related task that has already been performed. In Deep Learning, many problems are solved by ways of transfer learning. In computer vision for example tasks such as image classification, recognition or automatic caption all derive from similar architectures of deep networks. Models that are pre-trained on extensive data-sets such as Imagenet are constantly being re-used to solve new problems. The previous chapter offers a theoretical basis for transfer learning. We propose an approach where a model pre-trained on a given transport problem can be re-used to solve a "similar" problem. Reservoir models typically differ in their boundary conditions and permeability fields but all respond to a set of governing laws. A model that captures these physical laws can be transferred to many different problems. In our first attempt at solving the Buckley Leverett problem using only a set of boundary conditions, we were unsuccessful at training a network that would capture the behavior of the saturation. Instead, we used a model that was trained on a solution with given parameters and re-used some of the parameters of that solution to solve BL problems with different parameters/boundary conditions. This approach improves the solution and also converges faster as less parameters need to be trained. 

\begin{figure}%
    \centering
    \subfloat{\includegraphics[width=0.5\linewidth]{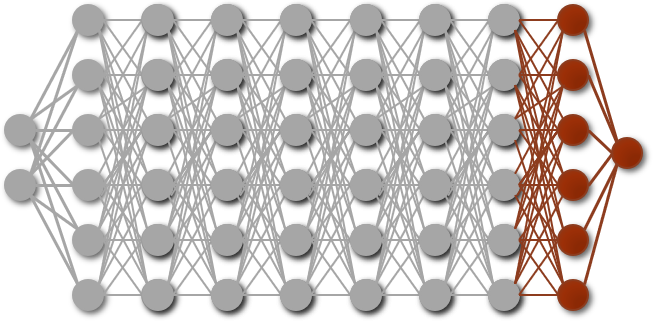} }%
    \qquad
    \subfloat{\includegraphics[width=1\linewidth]{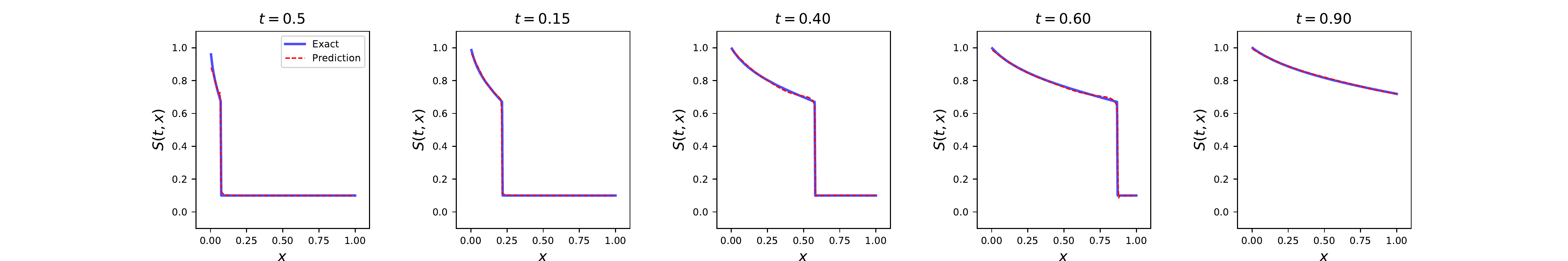} }%
    \caption{Results on inference case with transfer learning for cases with 6 frozen hidden layers}%
    \label{fig:BL_transfer_learning}%
\end{figure}

In our case, we pre-store the weights of a network that was trained to overfit a Buckley Leverett solution and re-train the network but only on the last few layers of the network. This approach allows to obtain a new solution without having to re-learn the general shape of the two-phase displacement. This approach leads to a better fit as shown in figure~\ref{fig:BL_transfer_learning}.

\section{Uncertainty Quantification, Variational Autoencoders and Generative Models}

\subsection{Introduction to Generative Models}
In the previous section, we have introduced neural networks that model the likelihood of a quantity of interest (QOI) such as water saturation knowing a set of observations and a physical constraint. These models are called \textbf{discriminative} as they try to learn mapping directly from the space of inputs ($x,t,S_{obs}$) and reject solutions that do not minimize the loss function. On the other hand, \textbf{generative} algorithms learn to model a distribution from which samples would maximize the likelihood estimation (or equivalently minimize the loss) and honor the constraints set in the learning problem. In other words, given a training data-set, generative models are capable of generating new samples from the same distribution. Several classes of generative models exist. They typically come in two categories:
\begin{itemize}
    \item Explicit density estimation in which an explicit form for the probability distribution of the model exists. Variational autoencoders are an exemple.
    \item Implicit density estimation in which we learn a model that can sample from the target density distribution without explicitly defining it. Generative adversarial networks are examples of this category
\end{itemize}
Examples of generative models include (but are not limited to):
\begin{itemize}
    \item Any parameterized probability distribution. A normal distribution $x\sim\mathcal{N}(\mu,\sigma^2)$.
    \item Gaussian Mixture Models where the type of distribution we sample from could be conditioned by a categorical variable $z_i$ $x_i|z_i=k\sim\mathcal{N}(\mu_k,\sigma_k^2)$. Note that the latent variable can be multi-dimensional.
    \item More modern examples include generative models for image generation. The model is trained on a large corpus of images of a certain class (hand written digits, faces, dogs, cats,...) and then tasked to generate new random images that look like the original ones. Variations in the latent variable then correspond to consistent changes in the generated image (pen stroke width, orientation, face expression). These have the advantage of not specifying the model or the latent variables. They are learned.
\end{itemize}
Random variable generation using a computer relies on a series of mathematical operations that are supposed to approximate the theoretical random process. Pseudo-random generators are able to construct sequences of numbers that approximately follow uniform random distribution while various techniques such as inverse transform method, Gibbs sampling, rejection sampling, Metropolis Hasting algorithm and other Markov Chain processes can turn uniform distribution into more complex ones.

Starting from any random variable $Z$, we learn a deterministic mapping that can transform the original distribution in any complex target distribution. The mapping is learned through a neural network function approximator. This leads to an output that is like (but not equal to) our observed data. This mapping allows the transformation of very simple random distributions (Uniform or Gaussian) into any type of distribution representing the data.

The inverse transform method is applied to generate random variables that follow a given distribution by "transformation" (inverse CDF) of a uniform random variable.
This principle can theoretically be applied to generate arbitrarily complex distributions. However, it requires knowledge of this transform function. Note that the probability space can feature several dimensions and non parametric distributions.

We use generative models to introduce noise or uncertainty in the quantity we try to simulate (water saturation at given time and position as an example). The GAN architecture for example allows for learning a function that will mimic the behavior of a front displacement without "learning" the data. This guarantees a greater stability to input variations and does not require transfer learning. It will also train robustly with a noisy input signal (whether it is the data-set or the parameters of the PDE).

Variational Autoencoders (or VAE), \cite{Kingma_2014} is an unsupervised learning method. It was popularized with generative models capable of reproducing complex distributions such as handwritten digits, faces and image parts. One of the building blocks of VAEs are {\bf generative models} which allow to randomly sample (or generate) new data points from a distribution similar to the observed one. This is accomplished by specifying a joint distribution between input, output and a latent variable. Latent variables represent encoded "concepts". One advantage of using latent variables is that they can serve to reduce the dimensionality of data.

\subsection{Generative Adversarial Networks}
The common application of Generative Adversarial Networks (GANS), \cite{goodfellow2014generative} is image generation \cite{berthelot2017began, karras2017progressive, ledig2017photo}, where the image of a face or a car is decomposed in $N$ dimensional vectors that belong to a certain distributions. Generating a new image of a car is equivalent to generating a random vector that follows the "car probability distribution" over the $N$ dimensional probability space. This corresponds to the sampling problem described earlier. The generation of such distributions from simple uniform ones becomes more challenging. In most cases, there may not even be an explicit form for this transformation. This is why neural networks can be used to model these complex distributions from data. The training of this network can be performed using a direct or indirect method. 

A direct training approach consists in comparing the true probability distribution with the generated one using a measure such as KL divergence or Maximum Mean Discrepancy (MMD). Backpropagation through the network allows for minimizing the distance measure.

Indirect training requires a downstream task over these two distributions. The generator's performance is now assessed with respect to this task. The downstream task of GANs is a discrimination task between true and generated samples. In a GAN architecture, we have a discriminator, that takes samples of true and generated data and tries to classify them as well as possible, and a generator that is trained to fool the discriminator as much as possible. The advantage of the indirect approach is that it provides a mechanism to produce several answers that are all valid but could differ from each other. Using a distance based approach like mean squared error tends to produce samples that resemble the average of the training set. This results in blurred out versions of the reality when applied to images. Indirect approach through the discriminator just learn to pass a sample if it looks real no matter how much it differs from others that do.

\cite{Yang2018PIGAN} presents an implicit variational inference formulation that constrains the generative model output to satisfy physical laws. This framework presents two advantages. This first one is that it does not require any form of transfer learning in order to obtain the correct solution. The second one is that it is suitable for characterizing uncertainty due to randomness of input or noise in observation. 

\subsection{Implementation and Results}
Our implementation of GAN for the two phase flow problem is inspired from \cite{Yang2018PIGAN_SPDE} and \cite{YangPerdikaris2018}. We cast the Transport problem into Two-player game. 
It follows the formulation of the problem as in \cite{raissi2017physicsI} (inference problem) with loss function expressed as
\begin{equation}
    \mathcal{L}_{PDE}(\theta) = \frac{1}{N_u}\sum\limits_{i=1}^{N_u}\|\hat{S}_{\theta}(\mathbf{x}_i,t_i) - \mathbf{S}_i\|^2 + \frac{1}{N_r}\sum\limits_{i=1}^{N_r}\|\mathcal{R}_{\theta}(\mathbf{x}_i,t_i)\|^2
\end{equation}
The novelty is to model uncertainty using a conditional latent variable models of the form:
\begin{eqnarray}
    p(\bold{S}|\bold{x},t) = \int p(\bold{S},\bold{z}|\bold{x},t)d\bold{z}\\
    = \int p(\bold{S}|\bold{x},t,\bold{z})p(\bold{z}|\bold{x},t)d\bold{z}
\end{eqnarray}
and constrain the resulting samples to honor the PDE.
\begin{equation}
    p(\bold{S}|\bold{x},t,\bold{z}), \bold{z} \sim p(\bold{z})
\end{equation}
Where $\bold{S}$ is solution to the PDE.

We train the generative network by matching the joint distribution of the generated samples with the joint distribution of the data. This is done by a minimization of the Kullback-Leibler divergence (KL).
\begin{equation}
    \mathbb{KL}[p_{\theta}(\bold{x},t,\bold{S})\|q(\bold{x},t,\bold{S})] = -h(p_{\theta}(\bold{x},t,\bold{S})) - \mathbb{E}_{p_{\theta}(\bold{x},t,\bold{S})}[log(q(\bold{x},t,\bold{S}))]
\end{equation}
Where $h(p_{\theta}(\bold{x},t,\bold{S}))$ is the entropy of the generative model. Maximizing the entropy promotes the spread of $p_{\theta}(\bold{x},t,\bold{S})$ while the second term encourages the supports of the two distribution to overlap. This mitigates mode collapse by introducing a regularization term. Mode collapse is when the generator generates a limited diversity of samples, or even the same sample, regardless of the input. We derive a lower bound for the entropy term:
\begin{equation}
    h(p_{\theta}(\bold{x},t,\bold{S})) \leq h(p(\bold{z})) + \mathbb{E}_{p_{\theta}(\bold{x},t,\bold{S},\bold{z})}[log(q_{\phi}(\bold{z}|\bold{x},t,\bold{S}))]
\end{equation}
Where $q_{\phi}(\bold{z}|\bold{x},t,\bold{S}))$ is a variational approximation to the true posterior over the latent variable.
The second term in the equation can be written:
\begin{eqnarray}
\mathbb{E}_{p_{\theta}(\bold{x},t,\bold{S})}[log(q(\bold{x},t,\bold{S}))] &= \displaystyle\int_{\mathbb{S_{p_{\theta}}}\cap \mathbb{S_{q}}}log(q(\bold{x},t,\bold{S}))p_{\theta}(\bold{x},t,\bold{u})d\bold{x}dtd\bold{S} + \\
&\displaystyle\int_{\mathbb{S_{p_{\theta}}}\cap \mathbb{S_{q}^o}}log(q(\bold{x},t,\bold{S}))p_{\theta}(\bold{x},t,\bold{u})d\bold{x}dtd\bold{S}
\end{eqnarray}
With $\mathbb{S_{p_{\theta}}}$ and $\mathbb{S_{q}}$ the support of the distribution for the associated functions and $\mathbb{S_{q}^o}$ the complimentary of $\mathbb{S_{q}}$. As opposed to \cite{Yang2018PIGAN}, we propose to approximate the two intractable integrals using two different networks $p_{\theta}$ (generator) and $q_{\phi}$ (posterior).

The discriminator emerges from the original definition of the KL divergence which is the difference between the cross entropy and entropy of the modeled probability function:

\begin{equation}
    \mathbb{KL}[p_{\theta}(\bold{x},t,\bold{S})\|q(\bold{x},t,\bold{S})] = \mathbb{E}_{p_{\theta}(\bold{x},t,\bold{S})}\left[\log\left(\frac{p_{\theta}(\bold{x},t,\bold{S})}{q(\bold{x},t,\bold{S})}\right)\right]
    \label{eq:KL_orig}
\end{equation}

The ratio of the two probability distribution functions is computed through a parametrization trick that introduces a classifier $\mathcal{D}_{\psi}$ trained to recognize the two density distributions. The density ratio written in eq.~\ref{eq:KL_orig} can be written:

\begin{align*}
    \frac{p_{\theta}(\bold{x},t,\bold{S})}{q(\bold{x},t,\bold{S})} &= \ddfrac{\rho(\bold{x},t,\bold{S}|y=1)}{\rho(\bold{x},t,\bold{S}|y=-1)}\\
    &= \ddfrac{\rho(y=1|\bold{x},t,\bold{S})\rho(\bold{x},t,\bold{S})\rho(y=-1)}{\rho(y=1)\rho(y=-1|\bold{x},t,\bold{S})\rho(\bold{x},t,\bold{S})}\\
    &= \ddfrac{\rho(y=1|\bold{x},t,\bold{S})}{1-\rho(y=1|\bold{x},t,\bold{S})}\\
    &= \ddfrac{\mathcal{D}_{\psi}(\bold{x},t,\bold{S})}{1 - \mathcal{D}_{\psi}(\bold{x},t,\bold{S})}
\end{align*}

The two coupled objective functions define the adversarial problem. We try to minimize the discriminator loss $\min_{\psi}\mathcal{L}_{\mathcal{D}}(\psi)$ and minimize a loss function that is a combination of generative loss and PDE loss $\min_{\theta,\phi}\mathcal{L}_{\mathcal{G}}(\theta,\phi)+\beta\mathcal{L}_{PDE}(\theta)$. These two losses are defined as:
\begin{equation}
    \mathcal{L}_{\mathcal{D}}(\psi) = - \mathbb{E}_{q(\bold{x},t,\bold{z})}[\log(1-\sigma(\mathcal{D}_{\psi}(\bold{x},t,\bold{S})))] -\mathbb{E}_{q(\bold{x},t)p(\bold{z})}[\log\sigma(\mathcal{D}_{\psi}(\bold{x},t,p_{\theta}(\bold{x},t,\bold{z})))]
\end{equation}
\begin{equation}
    \mathcal{L}_{\mathcal{G}}(\theta, \phi) = \mathbb{E}_{q(\bold{x},t)p(\bold{z})}[\mathcal{D}_{\psi}(\bold{x},t,p_{\theta}(\bold{x},t,\bold{z})) +(1-\lambda)\log(q_{\phi}(\bold{z}|\bold{x},t,p_{\theta}(\bold{x},t,\bold{z})))]
\end{equation}

The problem is cast as:
\begin{align}
    \min_{\psi} \mathcal{L}_{\mathcal{D}}(\psi)\\
    \min_{\theta,\phi}\mathcal{L}_{\mathcal{G}}(\theta,\phi) + \beta \mathcal{L}_{PDE}(\theta)
\end{align}

Similarly to \cite{Yang2018PIGAN} Three networks are jointly trained to address the minimax problem posed in GAN implementation:
\begin{itemize}
    \item Discriminator network $\mathcal{D}_{\psi}$: try to distinguish physical looking solutions from fake maximize objective s.t. $D(x)$ close to 1 (real) and $D(G(z))$ close to 0 (fake)
    \item Generator network $p_{\theta}$: try to fool the discriminator by generating physical looking solutions minimize objective such that $D(G(z))$ is close to 1 (discriminator is fooled)
    \item Posterior network $q_{\phi}$: variational approximation of the true data
\end{itemize}

\begin{figure}[H]
    \centering
    \includegraphics[width=1\linewidth]{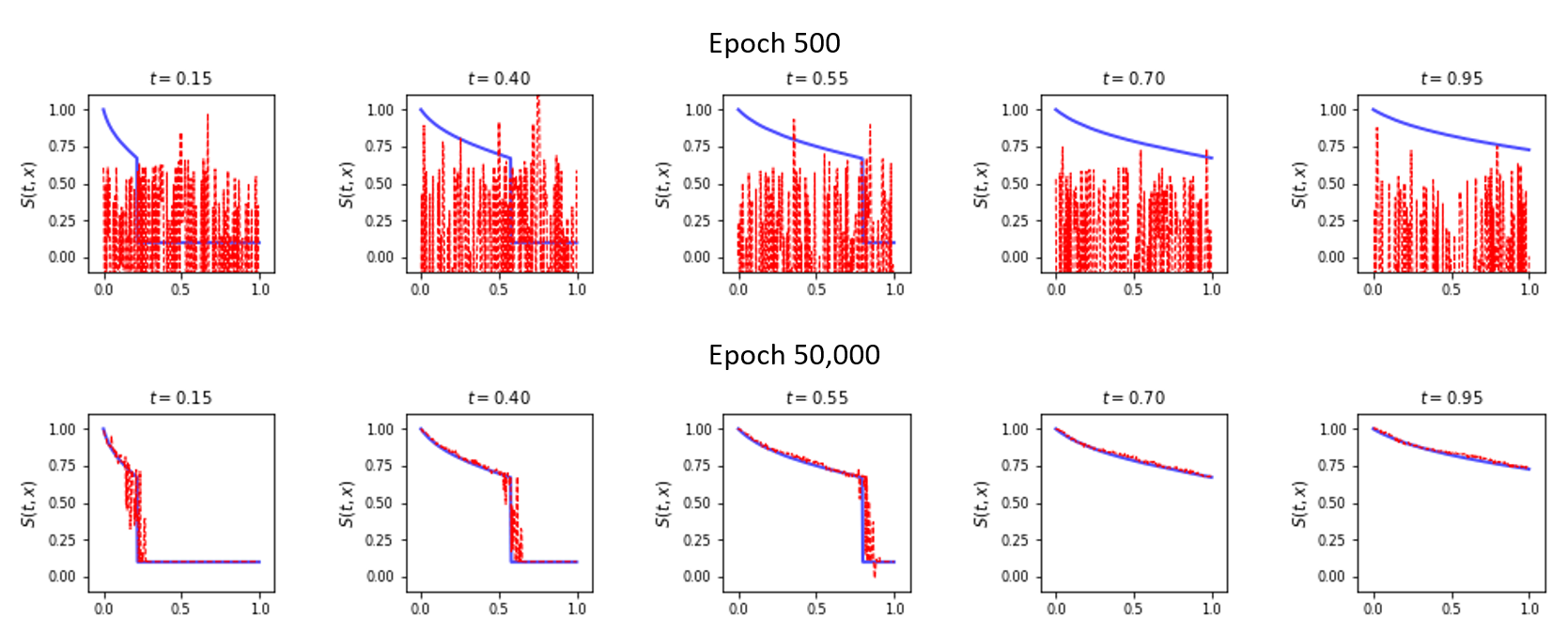}
    \caption{Training of PINNs using a generative adversarial network approach for two 1D transport problems. Horizontal Buckley Leverett at epoch 500 (top) and epoch 50,000 (bottom)}
    \label{fig:GAN_BL_training_example}
\end{figure}
We see that the GAN manage to simulate forms that are close to the actual ones. The models can also be trained on noisy data and this leads to a model that captures the uncertainty propagation and output distribution. 

\subsubsection{Diffusion}
\begin{figure}[H]
    \centering
    \includegraphics[width=1\linewidth]{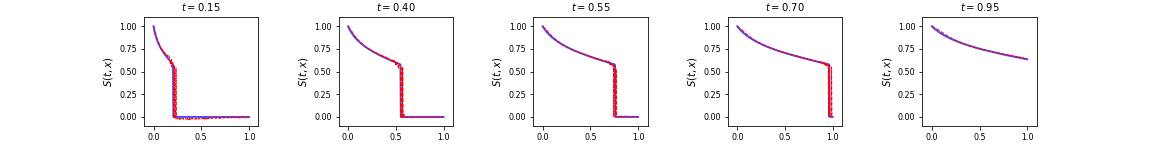}
    \caption{Training of PINNs using a generative adversarial network approach for two 1D transport problems with a diffusive term added to the PDE. Horizontal Buckley Leverett at epoch 15,000}
    \label{fig:GAN_BL_training_example_diffusion}
\end{figure}

We noted that training was greatly enhanced by adding a second order term to the original different partial differential equation. Eq.\ref{eq:Buckley_residual} becomes:
\begin{equation}
\label{eq:Buckley_residual_diff}
    \mathcal{R} = \frac{\partial \hat{S}}{\partial t} + \frac{\partial f}{\partial S}\frac{\partial \hat{S}}{\partial x} - \epsilon\frac{\partial^2 \hat{S}}{\partial x^2}
\end{equation}
This leads to a division of the number of epochs by two along with more accurate solutions as seen in figure~\ref{fig:GAN_BL_training_example_diffusion}
This result is quite intriguing and it is still unclear to us why the addition of this second term helps. A hypothesis is that like in classical methods of resolutions for hyperbolic problems of the sort, the weak formulation of the PDE along with boundary conditions does not provide full closure to the problem. When solving the hyperbolic problem using analytical methods (such as the method of characteristics), in the case where the characteristics cross, a shock forms and the flux velocities on each side of the shock must follow the entropy (or Oleinik) condition stipulating that upstream flux velocities should be higher than downstream ones. Otherwise the problem is not well posed. \cite{Lax_Hyperbolic} provides a detailed proof of this statement along with the explanation why in the case of a discrete resolution using a finite volume method, numerical diffusion allows to ignore the entropy condition.
This hypothesis is reinforced by results presented in section \ref{sec:gravity} where we are interested in the problem of gravity segregation where the added physics acts as another closure condition for the hyperbolic problem.

\subsubsection{Uncertainty Quantification}
\begin{figure}%
    \centering
    \includegraphics[width=1\linewidth]{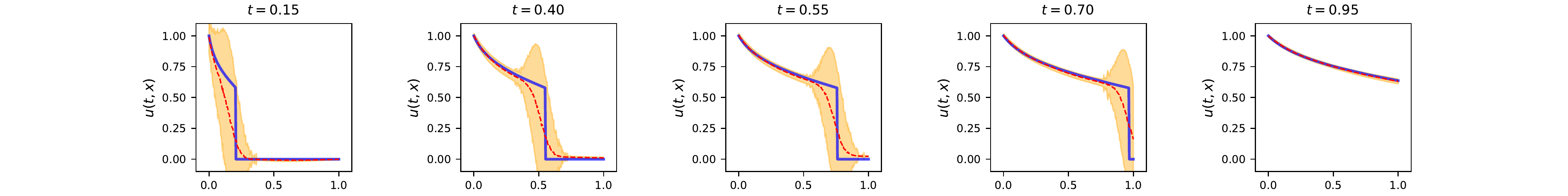}%
    \caption{Results on inference case with uncertainty propagation. The blue line is a representative solution, the red dotted line is the average over an ensemble of 1000 solutions and the beige enveloppe is the ensemble of solutions within the 2 standard deviation range from the average. This result is obtained after 15,000 training epochs.}%
    \label{fig:1d_results_noisy_init}%
\end{figure}

In this section, we are interested in evaluating the behavior of the GAN solution when the training data carries uncertainty. We are interested in the propagation of that uncertainty. We add a random variable $U\sim \mathcal{N}(\mu=0,\,\sigma^{2}=0.05)$ to the initial saturation condition and observe how it propagates through the system.
Figure~\ref{fig:boundary_noisy_input} shows the distribution of the initial saturation values and how it materializes as an initial condition for training:
\begin{figure}%
    \centering
    \subfloat{\includegraphics[width=0.45\linewidth]{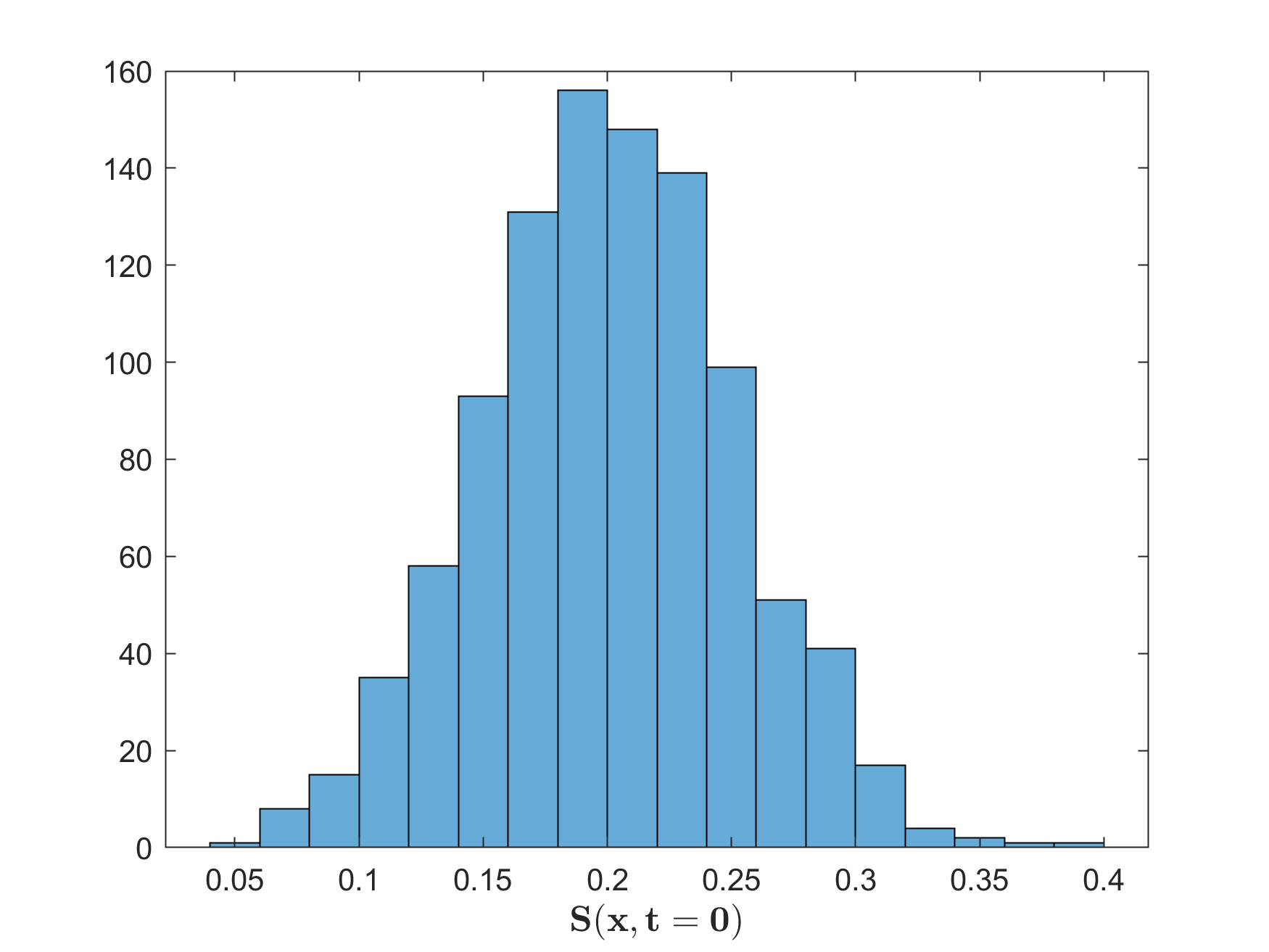}}
    \qquad
    \subfloat{\includegraphics[width=0.5\linewidth]{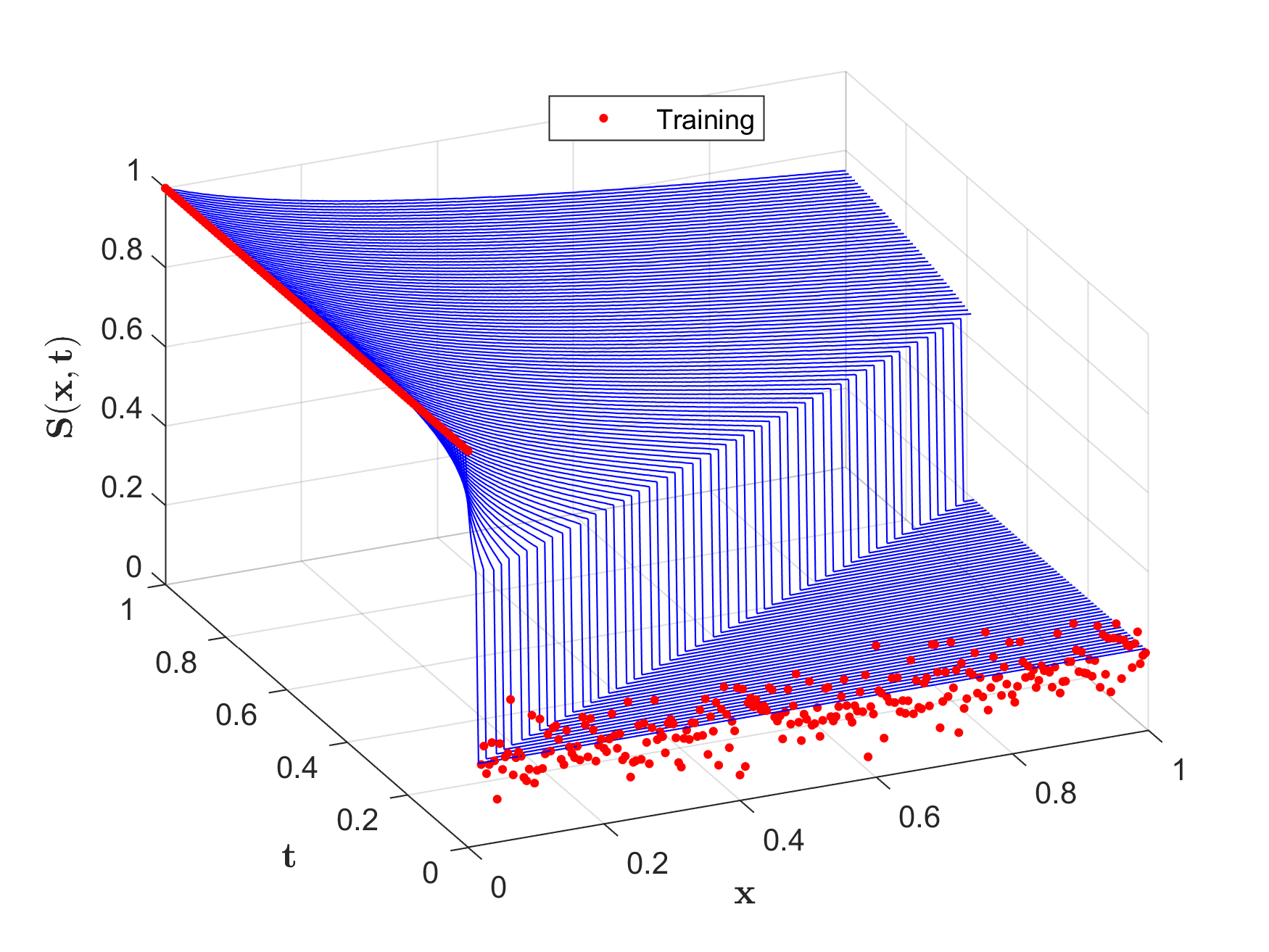}}
    \caption{Initial saturation distribution (left) and Training sample (right) for case with noisy input.}%
    \label{fig:boundary_noisy_input}%
\end{figure}
The model built with a GAN is capable of propagating the uncertain initial condition across time and produces saturation solutions that have a distribution associated. Figure~\ref{fig:1d_results_noisy_init} shows the range of solutions across time with the average solution at each time and an enveloppe representing two standard deviations away from the mean.
We notice that the region with highest uncertainty follows the shock in the solution. This is partially in line with what we would expect. Indeed, the propagation fails to capture the uncertainty downstream of the shock and we suspect that this is partially due to the number of collocation points chosen for the training. Indeed, with $10,000$ collocation points trained with a fixed parameter for the residual water saturation ($S_{wc}=0$) in eq.~\ref{eq:frac_flow}, the solution tends to force a saturation downstream of the shock to be close to the connate saturation regardless of the initial condition.
Another area of improvement is the sensitivity of the uncertainty envelope to the number of training epochs. Generally, once we have captured a representative solution on the training, increasing the number of epochs tends to reduce the uncertainty as the GAN will overfit the noise and produce a unique solution to the problem. A way to remedy this is to impose an early stopping condition to the training (We estimate that for the current architecture, 15,000 epochs is the right compromise between accurate solution and uncertainty quantification. Another solution could be to produce an ensemble of GAN solutions with different seeds and sample from the posterior in order to get the output distribution.

\subsection{Introducing Gravity}
\label{sec:gravity}
\begin{figure}[H]
    \centering
    \includegraphics[width=1\linewidth]{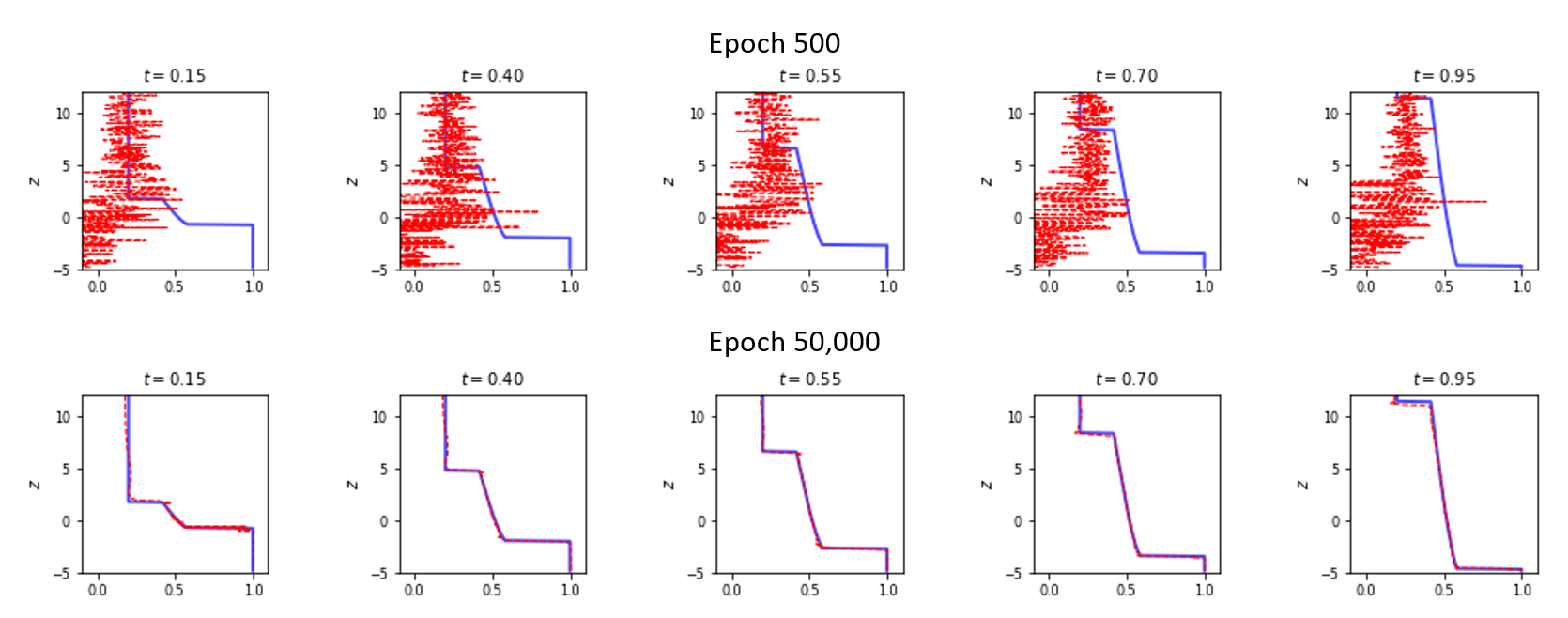}
    \caption{Training of PINNs using a generative adversarial network approach for a 1D transport problems. Gravity segregation of two phases at epoch 500 (top) and epoch 50,000 (bottom)}
    \label{fig:BL_solution_1D_gravity}
\end{figure}

When we introduce gravity to the two phase transport problem, the fractional flow curve takes the form:
\begin{equation}
    f_w(S_w,\theta) = \frac{\lambda_w}{\lambda_T}\left(1 - \frac{\lambda_o A}{Q}\Delta\rho g\sin\theta\right)
    \label{eq:BL_gravity}
\end{equation}
We refer to \cite{ERE221} for a detailed development of the analytical solution to this problem.
The fractional flow becomes a function of the dip angle $\theta$. The effects of that angle are the following:
\begin{enumerate}
    \item For flow updip ($\theta>0$), gravity delays the water flow
    \item For flow downdip ($\theta<0$), gravity enhances the water flow
\end{enumerate}
We define the gravity number, $N_g$ as the ratio of gravity to viscous forces:
\begin{equation}
    N_g = \frac{kA\Delta\rho g}{\mu_o Q}
\end{equation}
If the ratio of $Q/A$ is large, the flow is viscous dominated and gravity can be ignored. With this relationship and the Honarpour relative permeability relationship, we obtain an explicit formulation of the relative permeability:
\begin{equation}
    f_w(S_w,\theta) = \frac{1 - N_gk_{ro}^0(1 - S_w^*)^{n_o}\sin\theta}{1 + \frac{(1 - S_w^*)^{n_o}}{M^0(S_w^*)^{n_w}}}
\end{equation}
Where,
\begin{equation}
    S_w^* = \frac{S_w - S_{wc}}{1 - S_{or} - S_{wc}}
\end{equation}
Gravity has an effect on the shape of the fractional flow curve
\begin{figure}[H]
\centering
    \subfloat{\includegraphics[width=7cm]{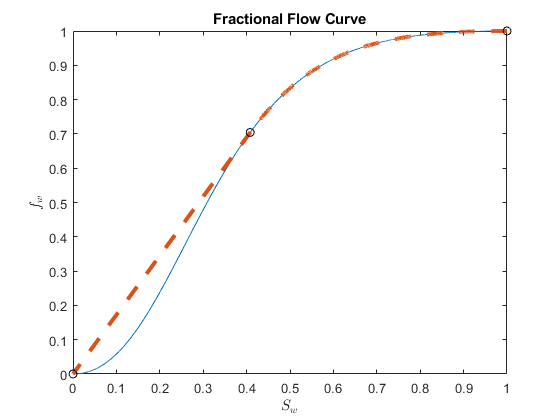} }%
    \qquad
    \subfloat{\includegraphics[width=7cm]{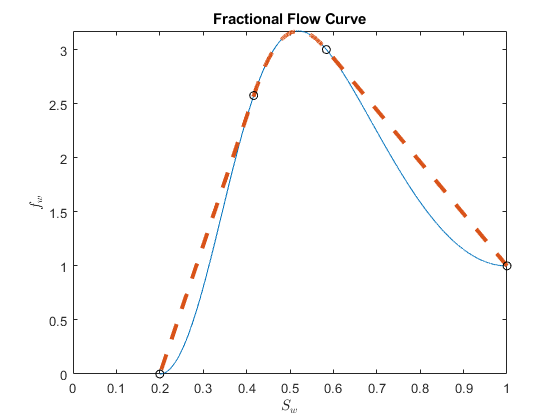} }%
   \caption{Fractional flow curve for horizontal Buckley-Leverett (left), gravity segregation with $N_g\sin\theta = -10$ and $M=5$ (right)}
\label{fig:frac_flow_gravity}
\end{figure}
The solution features two shocks with opposite velocities:

We train the system using the same approach as for the horizontal flow. The residual equation is updated (Eq.~\ref{eq:BL_gravity}) to reflect the presence of gravity. Results are shown in figure~\ref{fig:BL_solution_1D_gravity}

The predictions are of comparable quality and the architecture used is the same as for the previous problem (in the absence of gravity). This demonstrates the ability of the network to honor arbitrary physics and boundary/initial conditions. We also can observe in figure~\ref{fig:BL_solution_1D_gravity} that the solution with gravity is more accurate than the horisontal Buckley Leverett shown in figure~\ref{fig:GAN_BL_training_example}. This may be another manifestation of the effect observed with diffusion.

\subsection*{Discussion: Generalization to 2D and 3D simulation}
\begin{figure}[H]
\begin{center}
\includegraphics[width=1\linewidth]{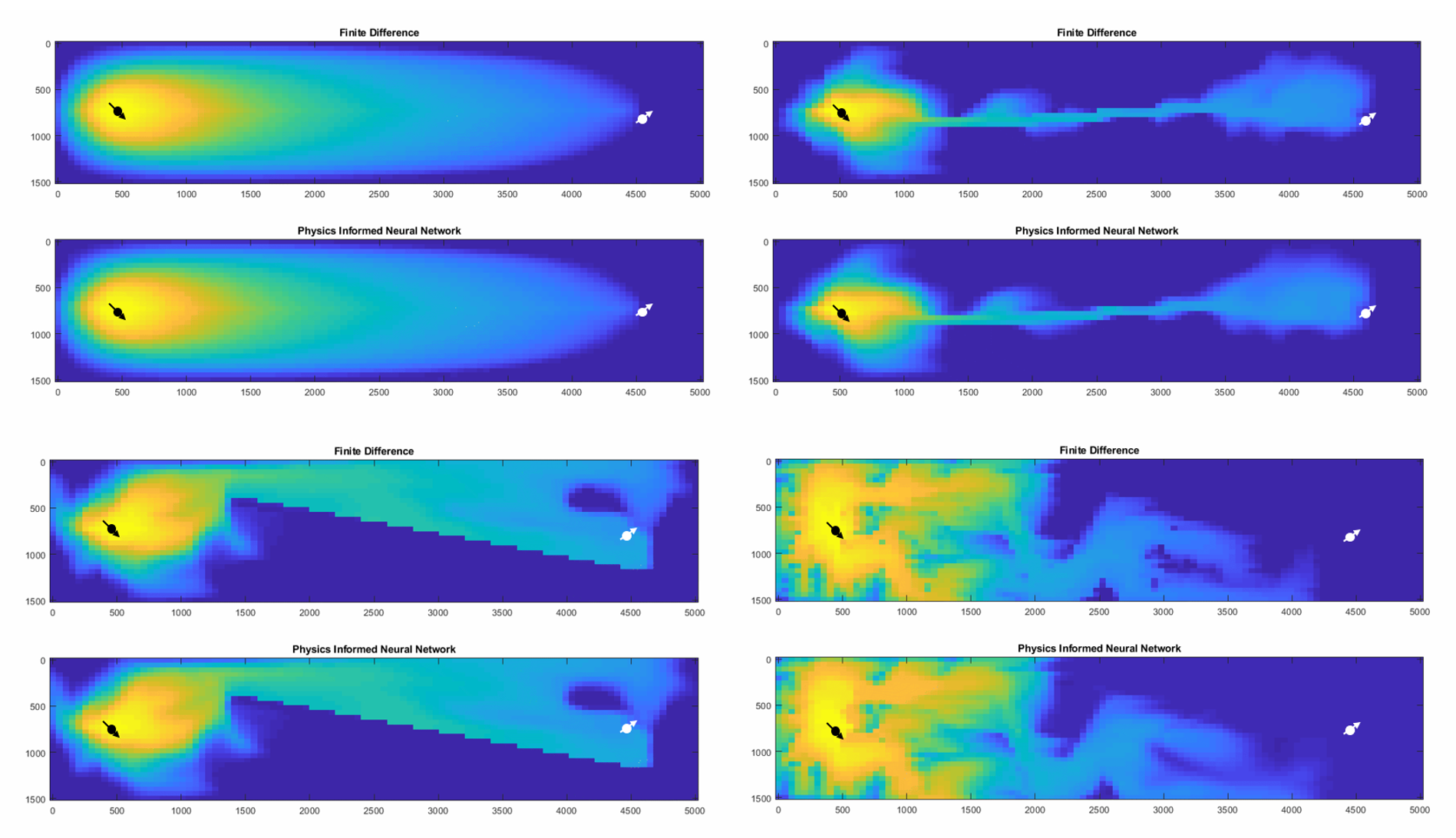}
\end{center}
   \caption{Comparison of 2D water/oil displacement cases in various permeability fields. The top left case is a homogeneous permeability field, top right is a case with a channel at the center of the field, bottom left is a case with a fault and bottom right is the top layer of SPE10.}
\label{fig:BL_solution_2D}
\end{figure}

One of the clearest advantages of this approach is its applicability to higher dimensional systems. This results from the decorrelation between information and resolution/computational complexity. The neural network encodes representations, independently of scale. This property makes the networks very attractive when it comes to resolving numerical problems that have poor complexity scaling (such as reservoir simulation). We apply very similar architecture (8 layers 20 neurons per layer) for 2D transport problems. We show 4 cases of 2D transport in figure~\ref{fig:BL_solution_2D}.

This method is a new way to produce high fidelity surrogates that honor governing laws. It can be perceived as a data assimilation under physics constraints for virtually any type of modeling. The computation times are order of magnitudes smaller for the neural network approach than they are for the finite volume approach and we suspect that the gain grows as the size and number of discrete elements gets larger. This work is under progress.

We show a new way to approach the problem of probabilistic surrogates that display high fidelity and honoring of governing laws. This work is under progress and we are still trying to answer how to truly generalize and explain the results we are getting. We have focused on coming up with a robust implementation of the forward problem knowing that all we do can be applied to data assimilation under physics constrained for virtually any type of modeling we can think of. Moreover, the latent variable we use on the GAN can be seen as an encoding of physical and interpretable low-dimensional features which would lead to a new, physically consistent dimensionality reduction technique for reservoir engineering.

The documentation and complete code for \verb+PI-GAN+ may be found at
\begin{center}
  \url{https://github.com/tenokonda/gan-pi.git}
\end{center}



\bibliographystyle{unsrt}  
\bibliography{references}  

\begin{thebibliography}{10}

\bibitem{Arps1945}
J.J. Arps.
\newblock {Analysis of Decline Curves}, 1945.

\bibitem{DeBezenac2017}
Emmanuel de~Bezenac, Arthur Pajot, and Patrick Gallinari.
\newblock Deep learning for physical processes: Incorporating prior scientific
  knowledge.
\newblock {\em CoRR}, abs/1711.07970, 2017.

\bibitem{Reichstein2019}
Bjorn Stevens Martin Jung Joachim Denzler Nuno~Carvalhais Markus~Reichstein,
  Gustau Camps-Valls and Prabhat.
\newblock Deep learning and process understanding for data-driven earth system
  science.
\newblock {\em Nature}, 2019.

\bibitem{BuckleyLeverett1942}
S.~E. Buckley and M.~C. Leverett.
\newblock Mechanism of fluid displacement in sands.
\newblock {\em Transactions of the AIME}, 146(01):107--116, 1942.

\bibitem{Mohaghegh2011}
S.D. Mohaghegh, O.~Grujic, S.~Zargari, and M.~Kalantari.
\newblock {SPE 143875 Modeling , History Matching , Forecasting and Analysis of
  Shale Reservoirs Performance Using Artificial Intelligence Top-Down ,
  Intelligent Reservoir Modeling for Shale Formations}.
\newblock {\em SPE Digital Energy Conference and Exhibition}, page~14, 2011.

\bibitem{Sun2018}
J~Sun, X~Ma, M~Kazi, and C~S~E Icon.
\newblock {SPE-190104-MS Comparison of Decline Curve Analysis DCA with
  Recursive Neural Networks RNN for Production Forecast of Multiple Wells}.
\newblock 2018.

\bibitem{Chiarmaonte2018}
M.~M. Chiaramonte and M.~Kiener.
\newblock Solving differential equations using neural networks.
\newblock {\em Unknown}, 2018.

\bibitem{Rudy2018}
Steven L.~Brunton Samuel~Rudy, Alessandro~Alla and J.~Nathan Kutz.
\newblock Data-driven identification of parametric partial differential
  equations.
\newblock {\em arXiv-1806.00732}, 2018.

\bibitem{MoZabaras2018}
Nicholas Zabaras Xiaoqing~Shi Shaoxing~Mo, Yinhao~Zhu and JichunWu.
\newblock Physics-constrained deep learning for high-dimensional surrogate
  modeling and uncertainty quantification without labeled data.
\newblock {\em Water Resources Research}, 2018.

\bibitem{YinhaoZabaras2018}
Nicholas~Zabaras Yinhao~Zhua.
\newblock Bayesian deep convolutional encoder-decoder networks for surrogate
  modeling and uncertainty quantification.
\newblock {\em Journal of Computational Physics}, 2018.

\bibitem{YinhaoZabaras2019}
Phaedon-Stelios Koutsourelakisb Paris~Perdikaris Yinhao~Zhua, Nicholas~Zabaras.
\newblock Deep convolutional encoder-decoder networks for uncertainty
  quantification of dynamic multiphase flow in heterogeneousmedia.
\newblock {\em Journal of Computational Physics}, 2019.

\bibitem{Long2018_pdenet}
Xianzhong Ma Bin~Dong Zichao~Long, Yiping~Lu.
\newblock Pde-net: Learning pdes from data.
\newblock {\em Proceedings of the 35 th International Conference on Machine
  Learning}, 2018.

\bibitem{RaissiJML2018}
Maziar Raissi.
\newblock Deep hidden physics models: Deep learning of nonlinear partial
  differential equations.
\newblock {\em Journal of Machine Learning Research 19}, 2018.

\bibitem{raissi2017physicsI}
Maziar Raissi, Paris Perdikaris, and George~Em Karniadakis.
\newblock Physics informed deep learning (part i): Data-driven solutions of
  nonlinear partial differential equations.
\newblock {\em arXiv preprint arXiv:1711.10561}, 2017.

\bibitem{raissi2017physicsII}
Maziar Raissi, Paris Perdikaris, and George~Em Karniadakis.
\newblock Physics informed deep learning (part ii): Data-driven discovery of
  nonlinear partial differential equations.
\newblock {\em arXiv preprint arXiv:1711.10566}, 2017.

\bibitem{tensorflow2015-whitepaper}
Mart\'{\i}n Abadi, Ashish Agarwal, Paul Barham, Eugene Brevdo, Zhifeng Chen,
  Craig Citro, Greg~S. Corrado, Andy Davis, Jeffrey Dean, Matthieu Devin,
  Sanjay Ghemawat, Ian Goodfellow, Andrew Harp, Geoffrey Irving, Michael Isard,
  Yangqing Jia, Rafal Jozefowicz, Lukasz Kaiser, Manjunath Kudlur, Josh
  Levenberg, Dandelion Man\'{e}, Rajat Monga, Sherry Moore, Derek Murray, Chris
  Olah, Mike Schuster, Jonathon Shlens, Benoit Steiner, Ilya Sutskever, Kunal
  Talwar, Paul Tucker, Vincent Vanhoucke, Vijay Vasudevan, Fernanda Vi\'{e}gas,
  Oriol Vinyals, Pete Warden, Martin Wattenberg, Martin Wicke, Yuan Yu, and
  Xiaoqiang Zheng.
\newblock {TensorFlow}: Large-scale machine learning on heterogeneous systems,
  2015.
\newblock Software available from tensorflow.org.

\bibitem{Baydin2018_AD}
Alexey Andreyevich Radul Jeffrey Mark~Siskind Atilim Gunes~Baydin, Barak
  A.~Pearlmutter.
\newblock Automatic differentiation in machine learning: a survey.
\newblock {\em arXiv}, 2018.

\bibitem{HORNIK1991251}
Kurt Hornik.
\newblock Approximation capabilities of multilayer feedforward networks.
\newblock {\em Neural Networks}, 4(2):251 -- 257, 1991.

\bibitem{NIPS2017_7203}
Zhou Lu, Hongming Pu, Feicheng Wang, Zhiqiang Hu, and Liwei Wang.
\newblock The expressive power of neural networks: A view from the width.
\newblock In I.~Guyon, U.~V. Luxburg, S.~Bengio, H.~Wallach, R.~Fergus,
  S.~Vishwanathan, and R.~Garnett, editors, {\em Advances in Neural Information
  Processing Systems 30}, pages 6231--6239. Curran Associates, Inc., 2017.

\bibitem{Henri_LipschitzReg}
Henry Gouk, Eibe Frank, Bernhard Pfahringer, and Michael Cree.
\newblock Regularisation of neural networks by enforcing lipschitz continuity.
\newblock 04 2018.

\bibitem{IntriguingNN2014}
Christian Szegedy, Wojciech Zaremba, Ilya Sutskever, Joan Bruna, Dumitru Erhan,
  Ian Goodfellow, and Rob Fergus.
\newblock Intriguing properties of neural networks.
\newblock In {\em International Conference on Learning Representations}, 2014.

\bibitem{DBLP_Shwartz-ZivT17}
Ravid Shwartz{-}Ziv and Naftali Tishby.
\newblock Opening the black box of deep neural networks via information.
\newblock {\em CoRR}, abs/1703.00810, 2017.

\bibitem{Shannon_1948}
C.~E. Shannon.
\newblock A mathematical theory of communication.
\newblock {\em The Bell System Technical Journal}, 27(1):379--423, 1948.

\bibitem{buckley1942mechanism}
Se~E Buckley, MCi Leverett, et~al.
\newblock Mechanism of fluid displacement in sands.
\newblock {\em Transactions of the AIME}, 146(01):107--116, 1942.

\bibitem{mohsen1985modification}
MFN Mohsen et~al.
\newblock Modification of welge's method of shock front location in the
  buckley-leverett problem for nonzero initial condition (includes associated
  papers 15193 and 15282 and 15797 and 15915 and 16458).
\newblock {\em Society of Petroleum Engineers Journal}, 25(04):521--523, 1985.

\bibitem{Kingma_2014}
Max~Welling Diederik P.~Kingma.
\newblock Auto-encoding variational bayes.
\newblock {\em arXiv}, 2014.

\bibitem{goodfellow2014generative}
Ian Goodfellow, Jean Pouget-Abadie, Mehdi Mirza, Bing Xu, David Warde-Farley,
  Sherjil Ozair, Aaron Courville, and Yoshua Bengio.
\newblock Generative adversarial nets.
\newblock In {\em Advances in neural information processing systems}, pages
  2672--2680, 2014.

\bibitem{berthelot2017began}
David Berthelot, Thomas Schumm, and Luke Metz.
\newblock Began: Boundary equilibrium generative adversarial networks.
\newblock {\em arXiv preprint arXiv:1703.10717}, 2017.

\bibitem{karras2017progressive}
Tero Karras, Timo Aila, Samuli Laine, and Jaakko Lehtinen.
\newblock Progressive growing of gans for improved quality, stability, and
  variation.
\newblock {\em arXiv preprint arXiv:1710.10196}, 2017.

\bibitem{ledig2017photo}
Christian Ledig, Lucas Theis, Ferenc Husz{\'a}r, Jose Caballero, Andrew
  Cunningham, Alejandro Acosta, Andrew Aitken, Alykhan Tejani, Johannes Totz,
  Zehan Wang, et~al.
\newblock Photo-realistic single image super-resolution using a generative
  adversarial network.
\newblock In {\em Proceedings of the IEEE conference on computer vision and
  pattern recognition}, pages 4681--4690, 2017.

\bibitem{Yang2018PIGAN}
Paris~Perdikaris Liu~Yang.
\newblock Physics-informed deep generative models.
\newblock {\em arXiv:1812.03511}, 2018.

\bibitem{Yang2018PIGAN_SPDE}
George Em~Karniadakis Liu~Yang, Dongkun~Zhang.
\newblock Physics-informed generative adversarial networks for stochastic
  differential equations.
\newblock {\em arXiv:1811.02033}, 2018.

\bibitem{YangPerdikaris2018}
Paris~Perdikaris Yibo~Yang.
\newblock Adversarial uncertainty quantification in physics-informed neural
  networks.
\newblock {\em Journal of Computational Physics}, November 2018.

\bibitem{Lax_Hyperbolic}
Peter~D. Lax.
\newblock Hyperbolic systems of conservation laws and the mathematical theory
  of shock waves.
\newblock {\em Courant Institute of Mathematical Science}, 11, 1989.

\bibitem{ERE221}
Hamdi Tchelepi.
\newblock Lecture notes on multiphase flow in porous media, January 2009.

\end{thebibliography}

\end{document}